\documentclass{article}

\usepackage{PRIMEarxiv}

\usepackage[utf8]{inputenc} % allow utf-8 input
\usepackage[T1]{fontenc}    % use 8-bit T1 fonts
\usepackage{url}            % simple URL typesetting
\usepackage{booktabs}       % professional-quality tables
\usepackage{amsfonts}       % blackboard math symbols
\usepackage{nicefrac}       % compact symbols for 1/2, etc.
\usepackage{microtype}      % microtypography
\usepackage{lipsum}
\usepackage{fancyhdr}       % header
\usepackage{graphicx}       % graphics
\graphicspath{{media/}}  
\usepackage{float}  
\usepackage{amsmath}
\usepackage{tikz}
\usepackage{svg}
\usepackage{algorithm}
\usepackage{algpseudocode}
\usepackage{ amssymb }
\usepackage{caption}
\usepackage{subcaption}
\newtheorem{theorem}{Theorem}

\title{SIDAR: Synthetic Image Dataset \\ for Alignment \& Restoration}

\author{
  Monika Kwiatkowski \\
  Computer Vision \& Remote Sensing \\
  Technische Universität Berlin \\
  \texttt{m.kwiatkowski@tu-berlin.de} 
   \And
  Simon Matern \\
  Computer Vision \& Remote Sensing \\
  Technische Universität Berlin \\
  \texttt{s.matern@tu-berlin.de} 
    \AND
  Olaf Hellwich \\
  Computer Vision \& Remote Sensing \\
  Technische Universität Berlin \\
  \texttt{olaf.hellwich@tu-berlin.de} 
}

\begin{document}
\maketitle

\begin{abstract}
    Image alignment and image restoration are classical computer vision tasks. However, there is still a lack of datasets that provide enough data to train and evaluate end-to-end deep learning models. Obtaining ground-truth data for image alignment requires sophisticated structure-from-motion methods or optical flow systems that often do not provide enough data variance, 
    i.e., typically providing a high number of image correspondences, while only introducing few changes of scenery within the underlying image sequences.
    Alternative approaches utilize random perspective distortions on existing image data. However, this only provides trivial distortions, lacking the complexity and variance of real-world scenarios. Instead, our proposed data augmentation helps to overcome the issue of data scarcity by using 3D rendering: images are added as textures onto a plane, then varying lighting conditions, shadows, and occlusions are added to the scene. The scene is rendered from multiple viewpoints, generating perspective distortions more consistent with real-world scenarios, with homographies closely resembling those of camera projections rather than randomized homographies. For each scene, we provide a sequence of distorted images with corresponding occlusion masks, homographies, and ground-truth labels. The resulting dataset can serve as a training and evaluation set for a multitude of tasks involving image alignment and artifact removal, such as deep homography estimation, dense image matching, 2D bundle adjustment, inpainting, shadow removal, denoising, content retrieval, and background subtraction. 
    Our data generation pipeline is customizable and can be applied to any existing dataset, serving as a data augmentation to further improve the feature learning of any existing method. 
\end{abstract}

% keywords can be removed

\section{Introduction}

Many classical computer vision tasks deal with the problem of image alignment and homography estimation. This usually requires the detection of sparse key points and the computation of correspondences across multiple images. In recent years, increasingly more methods have begun to use neural networks for feature extraction and matching of key points \cite{sarlin2020superglue,detone2018superpoint}. However, there is a lack of datasets that provide sufficient data and variety to train models. In order to train end-to-end image alignment models, large datasets of high-resolution images are necessary. Existing end-to-end deep learning methods, therefore, utilize datasets containing sparse image patches \cite{hpatches_2017_cvpr}, dense correspondences from structure-from-motion (SfM)  datasets \cite{MegaDepthLi18,schops2017multi} or optical flow datasets \cite{Butler:ECCV:2012}. Image patches only provide sparse correspondences between images, SfM datasets often lack a variety of scenes, and optical flow data has a high correlation between images with small displacements. 

A synthetic data generation for image alignment and restoration (SIDAR) is proposed. A planar object is generated, and a texture is added to it to simulate an artificial painting. Randomized geometric objects, lights, and cameras are added to the scene. By rendering these randomized scenes changing illumination, specular highlights, occlusions, and shadows are added to the original image. The data generation uses images from the WikiArt dataset \cite{danielczuk2019segmenting} to create a large variety of different content. However, the data generation can take any image dataset as input.
For each image, several distorted images are created with corresponding occlusion masks and pairwise homographies.
The datasets can be used for multiple objectives, such as homography estimation, image restoration, dense image matching, and robust feature learning. The rendering pipeline can be configured to generate specific artifacts or any combination of them. It can be used as a data augmentation to any existing datasets to further improve the performance and robustness of existing methods. 

\begin{figure}[h]
    \centering
    \includegraphics[width=0.8\linewidth]{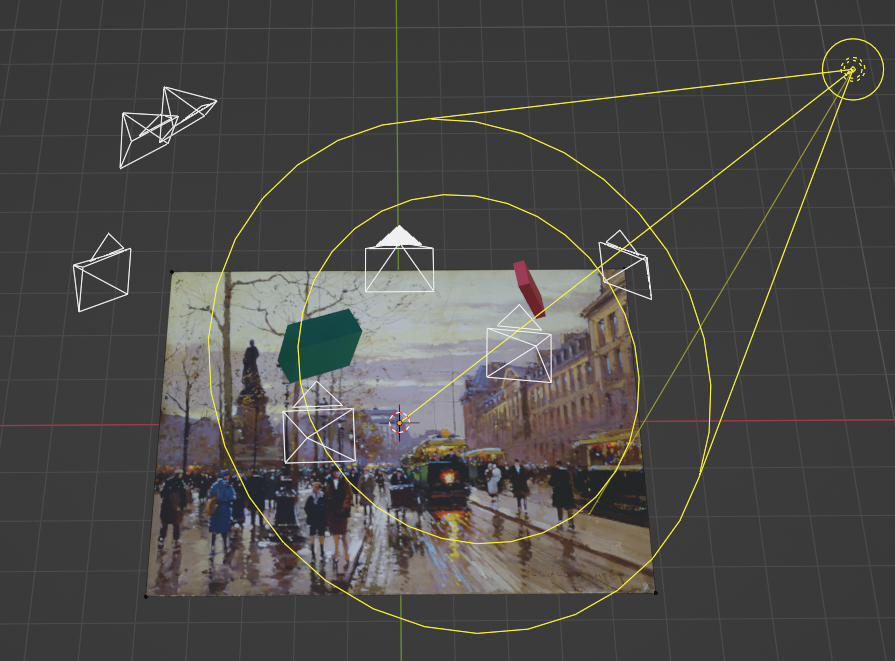}
    \caption{Illustration of a randomly generated scene using Blender. The plane shows a painting. The white pyramids describe randomly generated cameras; the yellow cone describes a spotlight. Geometric objects serve as occlusions and cast shadows onto the plane.}
    \label{fig:blender}
\end{figure}

Our dataset and rendering pipeline provide the following contribution:
\begin{itemize}
    \item Data quantity: An arbitrary number of distortions can be generated for each input image. Arbitrarily many scenes can be generated with an arbitrarily long image sequence each.
    \item Data variety: The input images and the image distortions can create a large variation in content and artifacts.
    \item Customizability: The rendering pipeline can be configured to control the amount and type of artifacts. For example, custom datasets can be created that only contain shadows or occlusions.
    \item Extendibility: The rendering pipeline can be extended to other datasets. It is also possible to change the rendering pipeline in order to generate new artifacts and data modalities, such as videos of moving objects.
\end{itemize}

One dataset is generated that only consists of aligned image sequences with various distortions, such as changing illumination, shadows, and occlusions. Another dataset is generated that, in addition to the aforementioned distortions, also contains perspective distortions. In order to reconstruct the underlying signal from a sequence of distortions, the information needs to be aligned first.  Shadows and illumination distort the original signal, but they do not remove the information completely. Occlusions cover the underlying content in parts of the image. However, for all of these cases, using additional images allows us to identify and remove distortions using pixel-wise or patchwise comparisons along the temporal dimension. Perspective distortions create misalignment that makes reconstruction more complicated. For this reason, we treat misaligned images as a categorically different problem. 

In the following chapters, we discuss related datasets and compare their advantages and shortcomings. The rendering pipeline is described in detail. Finally, the generated datasets are presented, and several use cases for end-to-end learning are discussed.
\section{Related Work}

In this section, we review some existing datasets and compare them with our dataset. We discuss the application of our dataset to various tasks.

\subsection{Image Restoration \& Background Subtraction}

Our data generation allows the creation of image sequences containing various distortions. In the case of a static camera, this is similar to a task such as background subtraction or change detection. Existing dataset for background subtraction use videos of a static scene \cite{vacavant2013benchmark, goyette2012changedetection,jodoin2017extensive,toyama1999wallflower,kalsotra2019comprehensive}. Each individual frame can contain deviations from an ideal background image. The variance can be due to changes in the background, such as illumination changes, weather conditions, or small movement of objects.  Additionally, distortions can be caused by foreground objects or camera noise. The goal is to extract a background model of the scene, which can be further used for segmentation into background and foreground. 

When training deep learning models, one can differentiate between scene-dependent models and scene-independent models \cite{mandal2021empirical}. Scene-dependent models learn a background model on a specific scene and must generalize on new images of the same scene, whereas scene-independent models learn are evaluated on new scenes.
Many existing datasets, such as CDNet \cite{goyette2012changedetection} or SBMNet\cite{jodoin2017extensive}, contain many individual frames but only a few different scenes. Training a model end-to-end on these datasets allows a robust training of scene-dependent models, but the low variation in scenes limits the generalization of models. 
Our data generation can create arbitrarily many variations within a specific scene and across different scenes. This can be useful to train and evaluate scene-independent models. Compared to video sequences, SIDAR has much more variation between each image, and there is no correlation between each frame. Methods that rely on optical flow cannot be used here.

The proposed SIDAR dataset can be customized to generate specific artifacts, such as changes in illumination, occlusions, and shadows. It can also create any combination of these artifacts. One can apply our dataset to train models for various tasks, such as the detection and removal of shadows, specular highlights, or occlusions. Existing datasets containing shadows \cite{wang2017stacked, kligler2018document,qu2017deshadownet} or illumination changes \cite{butler2012naturalistic,roberts:2021} often deal with each artifact individually and provide less control over their combinations. 

The main shortcoming of our dataset compared to existing datasets are that it is limited to a planar background and relies on synthetic data generation. Certain artifacts caused by the geometry of the scene or due to noise in the image formation from real cameras cannot be synthetically replicated.

\subsection{Homography Estimation}
\label{sec:homography-estimation}
Homography estimation describes a fundamental task in computer vision. However, there are few datasets providing enough ground truth homographies between image pairs to train a deep-learning model. A common approach to generate labeled data is to use any image dataset, such as MS COCO \cite{lin2014microsoft}, and apply perspective distortion \cite{chang2017clkn,detone2016deep,erlik2017homography}. These methods do not introduce any additional challenges or artifacts. Other methods apply perspective distortion to video sequences with static cameras \cite{Le_2020_CVPR,Cao_2022_CVPR}. The video sequences introduce dynamic objects that can create outliers when computing correspondences between images. In both cases, the perspective distortion is created artificially and does not follow a realistic planar projection. 

Datasets that rely on structure-from-motion or otherwise estimate homographies from real images, such as \textit{HPatches} \cite{hpatches_2017_cvpr}, \textit{Oxford Affine} \cite{mikolajczyk2005performance} or \textit{AdelaideRMF}\cite{wong2011dynamic} only provide enough image data to train deep learning models on smaller image patches for feature detection. 
\textit{NYU-VP} and \textit{YUD+} use detected planar objects in 3D scenes \cite{kluger2020consac} to develop a large dataset that is used for self-supervised training. \textit{NYU-VP} and \textit{YUD+} only provide sparse correspondences of line segments.

For all of these datasets, there is little variation in scenes, and each individual image does not contain many distortions. The proposed SIDAR dataset tries to overcome these shortcomings by generating strong distortions within each scene. Homographies between images can be computed regardless of the complexity of the scene and the presence of artifacts. As described in section \ref{sec:homography}, a homography can be computed between any image pair if the relative position of cameras and plane are known. Furthermore, we provide homographies between all images. This provides a unique setting where the relative orientation of all images can be jointly estimated under various distortions. It is possible to evaluate Bundle Adjustment methods under perturbations and develop end-to-end deep learning methods \cite{lin2021barf,lindenberger2021pixsfm}.

\subsection{Descriptor Learning \& Dense Correspondences}

Keypoint detection and image descriptors are fundamental methods in many computer vision tasks. Structure-from-motion and other photogrammetric methods rely on point correspondences computed from sparse key points \cite{hartley2003multiple}. A requirement for local feature detectors is to identify the location of distinct image points and compute a robust feature representation. The features should be invariant to various distortions, such as noise, changing illumination, scale, and perspective distortions. Recent developments in feature detectors use a data-driven approach to learn robust feature representations with deep learning \cite{NIPS2017_831caa1b,detone2018superpoint,shen2020ransac}. The descriptors are either trained on sparse image correspondences from structure-from-motion methods \cite{MegaDepthLi18} or by using perspective transformations on any image dataset, such as MS COCO \cite{lin2014microsoft}, as described in section \ref{sec:homography-estimation}. 

SIDAR provides the ground truth homographies between image pairs with arbitrarily many artifacts. The dataset allows explicitly adding illumination changes, shadows, specular highlights, etc., as data augmentations to train more robust descriptors. The homographies also provide dense correspondences for each pixel. This allows us to train models for dense image matching \cite{GLUNet_Truong_2020,pdcnet+}. Structure-from-motion datasets \cite{MegaDepthLi18,schops2017multi} also provide dense correspondences, but they often contain a limited amount of scenes, distortion, and only a few occlusions. By changing the texture of the image plane, SIDAR has a larger variety of patches and key points.

\section{Rendering Pipeline}
\label{chap:render}
Figure \ref{fig:blender} illustrates the data generation process. We use paintings from the Wiki Art dataset \cite{danielczuk2019segmenting} as our ground-truth labels. Any other image dataset could also be used, but Wiki Art contains a large variety of artworks from various periods and art styles. We believe that the diversity of paintings makes the reconstruction more challenging and reduces biases towards a specific type of image. We take an image from the dataset and use it as a texture on a plane in 3D.

Furthermore, we generate geometric objects and put them approximately in between the plane and the camera's positions. We utilize Blender's ability to apply different materials to textures. We apply randomized materials to the image texture and occluding objects. The appearance of an occluding object can be diffuse, shiny, reflective, and transparent. The material properties also change the effect lighting has on the plane.
It changes the appearance of specularities, shadows, and overall brightness. \\
Finally, we iterate over the cameras and render the images. Blender's physically-based path tracer, Cycles, is used for rendering the final image. Path tracing enables more realistic effects compared to rasterization. It allows the simulation of effects, such as reflections, retractions, and soft shadows. 
\subsection{Virtual Painting}

We first generate a 2D image plane. The plane lies on the $xy$-plane, i.e., the plane is described as:
\begin{align}
    0\cdot x +0 \cdot y + z = 0
\end{align}
 The center of the plane also lies exactly in the origin $(0,0,0)$.
The plane is described by its four corner points. Let $w,h$ be the width and height of the plane, then the corners are defined as:
\begin{align}
    \mathbf{X}_1 = \begin{pmatrix}
         w/2 \\
         h/2 \\
         0
    \end{pmatrix},
    \mathbf{X}_2 = \begin{pmatrix}
         -w/2 \\
         h/2\\
         0
    \end{pmatrix},
    \mathbf{X}_3 = \begin{pmatrix}
         w/2 \\
         -h/2\\
         0
    \end{pmatrix},
    \mathbf{X}_4 = \begin{pmatrix}
         -w/2 \\
         -h/2\\
         0
    \end{pmatrix}
\end{align}
We scale the plane along the $x$ and $y$ direction to fit the image's aspect ratio. We apply the given image as a texture to the plane.

\subsection{Aligned Camera}
\label{chap:alignment}
\begin{figure}[h]
    \centering
\includegraphics[width=\linewidth]{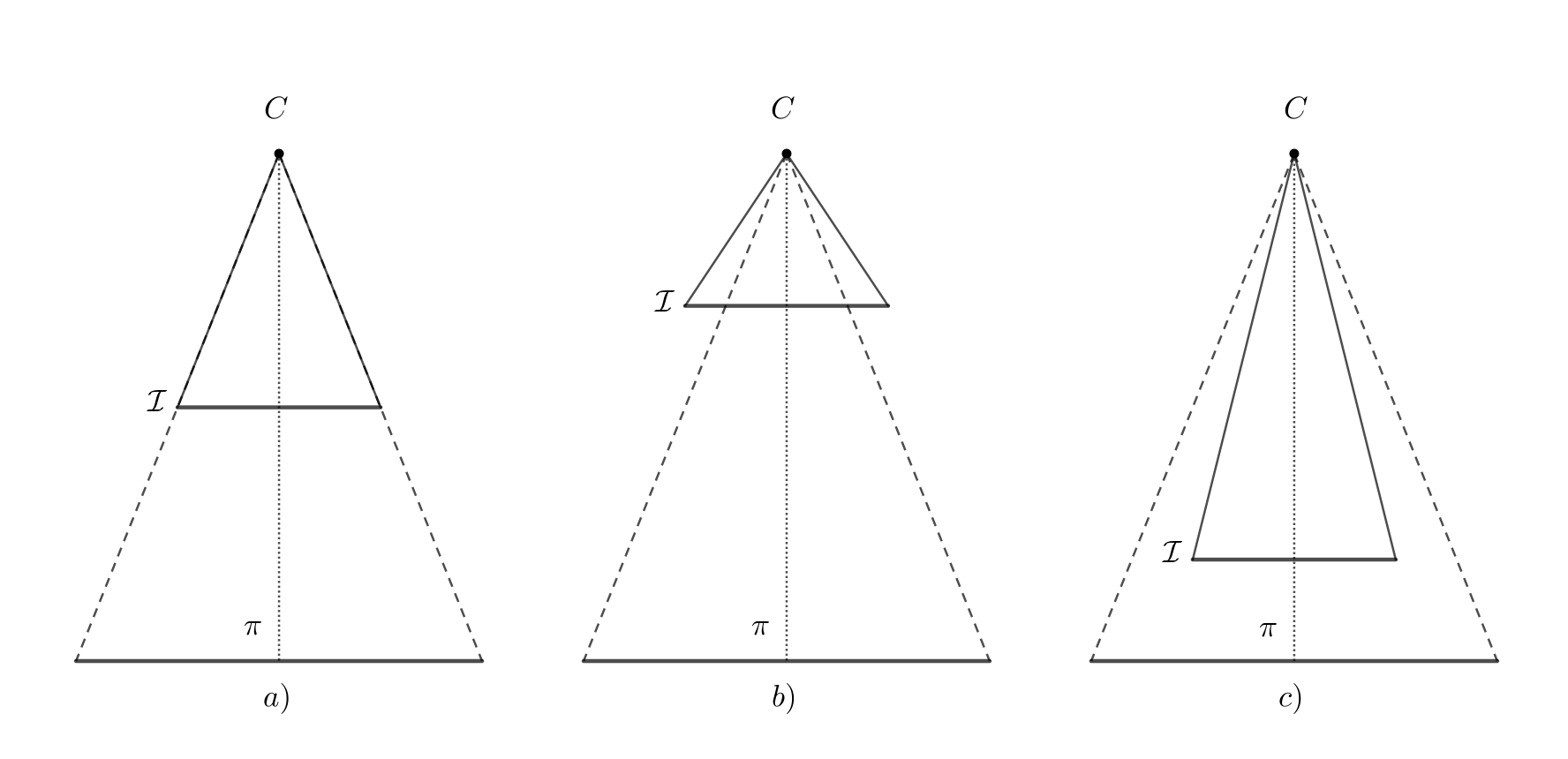}
    \caption{An Illustration of changing the principal distance between the projection center $C$ and the image plane $\mathcal{I}$. The position of the plane $\pi$ and the projection center are fixated. In a) the image plane $\mathcal{I}$ captures all of the content from the painting plane $\pi$. }
    \label{fig:alignment}
\end{figure}

To render the scene, we add virtual cameras to the scene. We differentiate between a camera that is aligned with the painting's plane and a setup that adds perspective distortions.
\label{seq:aligned}
 To enforce the alignment, we use a single static camera that perfectly fits the image plane. The camera's viewing direction is set to be perpendicular to the image plane and centered on the image plane. Our goal is to adjust the vertical and horizontal field of view such that only the image can be seen. Figure \ref{fig:alignment} illustrates this problem. 

We fixate on the camera's projection center, and we also fixate on the size of the sensor. We set the resolution and aspect ratio of the sensor equal to the painting. 
As can be seen in figure \ref{fig:alignment}, the alignment only depends on the principal distance. 
\begin{figure}[h]
    \centering
\includegraphics[width=\linewidth]{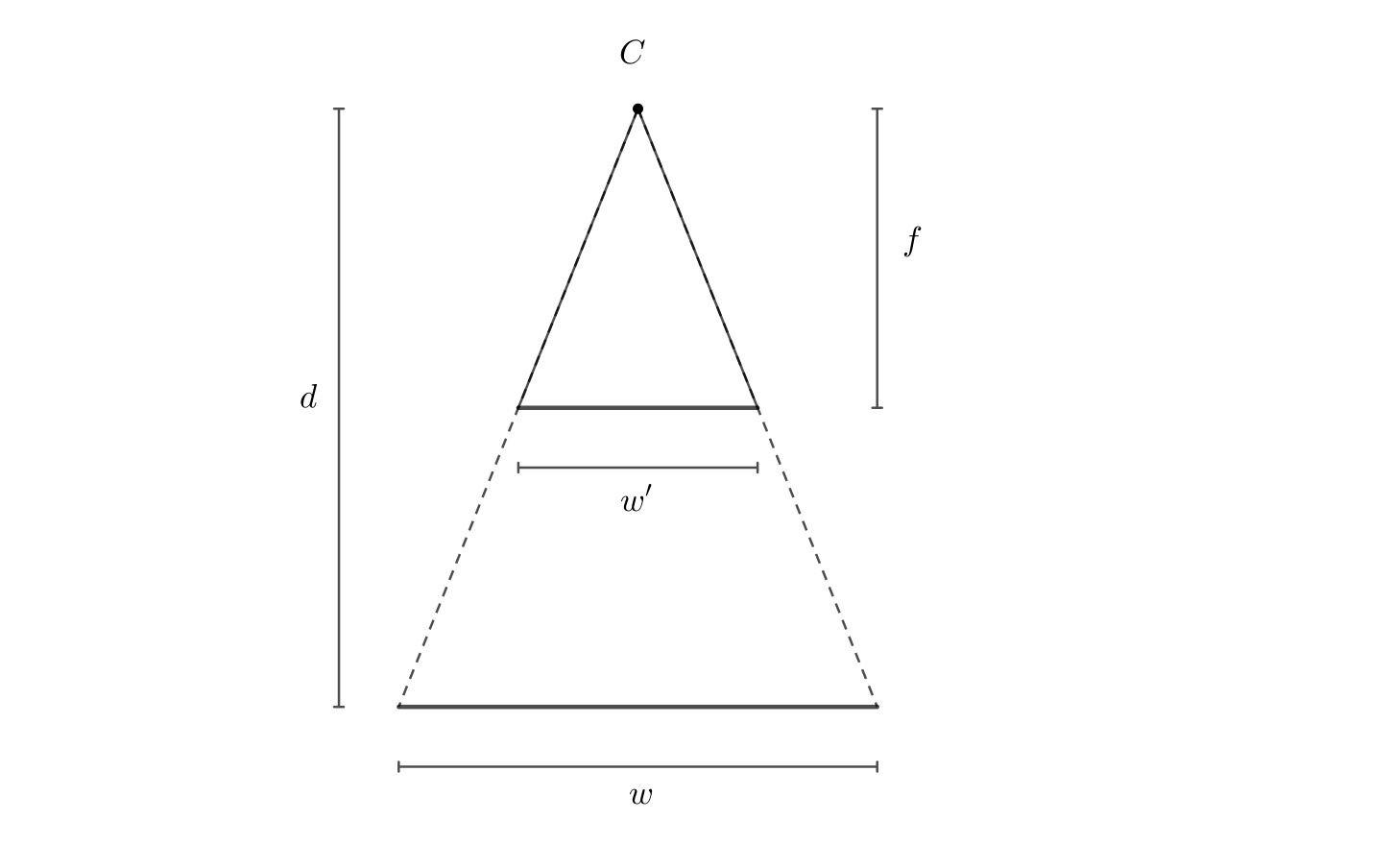}
    \caption{Illustration of a camera sensor with width $w'$ that is aligned with the image with width $w$. The principal distance is given as $f$, and the distance between the camera and image is given as $d$.}
    \label{fig:alignment-indiv}
\end{figure}

Let $f$ be the principal distance, $w$ the image width, $w'$ the sensor width, and $d$ the distance between the projection center and image as illustrated in figure \ref{fig:alignment-indiv}. The optimal $f$ can be computed from the intercept theorem:
\begin{align}
    \dfrac{f}{d} &= \dfrac{w'}{w} \\
    \Leftrightarrow f &= d\dfrac{w'}{w} 
\end{align}

Note that the previous formulas and graphics only describe the alignment with regard to a single dimension, i.e., image width. Let $h,h'$ be the image height and sensor height, respectively.   
\begin{align}
    \dfrac{w}{h} &= \dfrac{w'}{h'} \\
    \Leftrightarrow \dfrac{h'}{h} &= \dfrac{w'}{w} 
\end{align}
We can use the intercept theorem to find the optimal focal length $f'$ along the other dimension:
\begin{align}
    \dfrac{f'}{d} &= \dfrac{h'}{h} \\
    \Leftrightarrow f' &= d\dfrac{h'}{h}  = d\dfrac{w'}{w} = f \\
    \Rightarrow f' &= f
\end{align}
Since we set the sensor's aspect ratio equal to the image's aspect ratio, both dimensions share the same optimal focal length.

\subsection{Perspective Distortions}
\label{sec:homography}
\begin{figure}[H]
    \centering
    \includegraphics[width=0.8\linewidth]{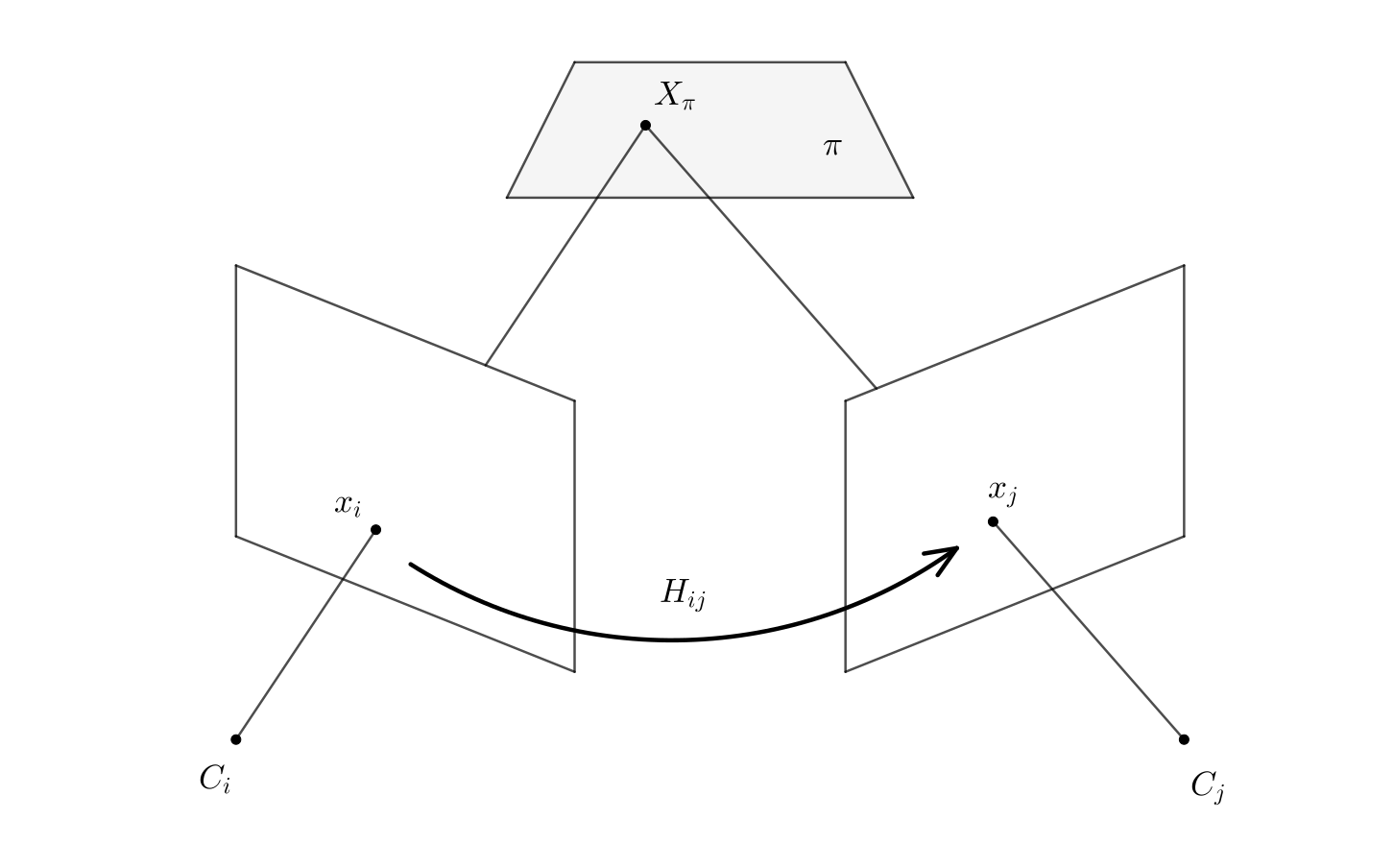}
    \caption{Illustration of a homography induced by a plane.}
    \label{fig:plane-homography}
\end{figure}

To create perspective distortions, we randomize the generation of cameras by sampling from a range of 3D positions. The field of view of the camera is also randomly sampled to create varying zoom effects. This also creates a variety of intrinsic camera parameters. Additionally, we center the viewing direction of the camera into the image center. This is done to guarantee that the image is seen by the camera. Otherwise, the camera might only see empty space. \\
Since we project a planar object onto image planes, all images are related by 2D homographies. This relationship is illustrated in figure \ref{fig:plane-homography}.
A point with pixel coordinates $(x,y)$ in image $i$ is projected onto the coordinates $(x',y')$ in image $j$ with the homography $H_{ij}$:
\begin{align}
    \lambda \begin{pmatrix}
    x' \\ y' \\ 1
    \end{pmatrix} = H_{ij}
    \begin{pmatrix}
    x \\ y \\ 1
    \end{pmatrix}
\end{align}

The following chapters explain two different methods for calculating the homography. Unlike in real datasets, these methods can always compute the corresponding homographies regardless of overlap and distortions. Real datasets rely on more complicated photogrammetric methods that try to find point correspondences between images and optimize the orientation of cameras using bundle adjustment \cite{hpatches_2017_cvpr,hartley2003multiple,MegaDepthLi18}.
Regardless of the complexity of the scene, the homographies can be precisely computed.

\subsubsection{Homography from Projection Matrices and Plane}
We can derive the homography by projecting the point $x_i$ onto a point on the plane $X_\pi$ and then project it into the second image to derive $x_j$ \cite{hartley2003multiple}. The resulting transformation describes the homography between both cameras.
We assume that the origin lies in the projection center of the first camera $C_i$.
\begin{theorem}
    Given two projection matrices $P_i = [I|0]$, $P_j = [A|a]$ and a plane defined as $\pi^TX =0$ with homogeneous coordinates $\pi = (v^T,1)^T$, the homography induced by the plane is:
    \begin{align}
        H_{ij} = A - av^T
    \end{align}
\end{theorem}

We can extend the previous theorem to a calibrated setting. 

\begin{theorem}
    Given two calibrated cameras described by projection matrices $P_i = K_i[I|0]$, $P_j = K_j[R|t]$ and a plane $\pi_E=(n^T, d)^T$, the homography induced by the plane is:
    \begin{align}
        H_{ij} = K_j\left(R - \dfrac{1}{d} tn^T\right)K_i^{-1} \label{eq:cal-homography}
    \end{align}
\end{theorem}

In order to compute the projective transformation between two arbitrary cameras, the relative orientation $R,t$ has to be extracted first by changing the origin into the projection center of the first camera. Furthermore, the plane has to be computed with regard to the new coordinate system.

\subsubsection{Direct Linear Transform}
\label{sec:dlt}
The homographies can be calculated using \eqref{eq:cal-homography} with Blender's scene graph. However, we decided to apply the Direct Linear Transform to point correspondences\cite{hartley2003multiple} since it does not require traversing the scene graph and transforming the objects relative to the new coordinate system. We can compute the homographies from the projection matrices and correspondences directly. \\
Homographies can be estimated from four point-correspondences:
\begin{theorem}
    Let $\mathbf{x}_i' = (x_i,y_i,w_i)^T$ and $\mathbf{x}_i' = (x_i,y_i,w_i)^T$ be the homogeneous coordinates of two corresponding points $\mathbf{x}_i \leftrightarrow \mathbf{x}_i'$. The points are related by a homography $ \mathbf{x}_i' = H\mathbf{x}_i$. Then the correspondence can be described by an under-determined homogeneous equation:
    \begin{align*}       
        A_ih = \begin{pmatrix}
            0 & -w_i'\mathbf{x}_i^T & y_i'\mathbf{x}_i^T \\
            -w_i'\mathbf{x}_i^T & 0 & x_i'\mathbf{x}_i^T
        \end{pmatrix}
        \begin{pmatrix}
            h_1 \\
            \vdots \\
            h_9
        \end{pmatrix} &= 0 \\
    \end{align*}
    where $A_i$ is a design matrix containing the measured point correspondences $\mathbf{x}_i \leftrightarrow \mathbf{x}_i'$ and $h=vec(H)$ is vectorized version of $H$.
    Using at least four correspondences, the homography can be estimated by concatenating into one larger homogeneous equation system:
    \begin{align}       
        Ah = \begin{pmatrix}
            A_1 \\
            \vdots\\
            A_4
        \end{pmatrix}
        \begin{pmatrix}
            h_1 \\
            \vdots \\
            h_9
        \end{pmatrix} &= 0 \label{eq:dlt} 
    \end{align}
    $h$ can be exactly determined by finding the null space of $A$.  
\end{theorem}
Since Blender provides the camera's positions in space and their intrinsic parameters, we can calculate the projection points explicitly.
 We project each of the painting's corners $X_k,~~k=1,..,4$ into the $i$-th image plane $\mathbf{x}_i^{(k)}$ using their projection matrices:
\begin{align}
    \mathbf{x}_i^{(k)} &= P_i X_k \\
   &= K_i [I|0]  \begin{bmatrix}
    R_i & t_i\\ 0 & 1
    \end{bmatrix} X_k ~~~~k=1,..,4
\end{align}
$(R_i, t_i)$ describes the global rotation and translation of the camera, $K_i$ describes the intrinsic camera parameters. Using the four-point pairs $\mathbf{x}_k^{(i)} \leftrightarrow \mathbf{x}_k^{(j)} ~~k=1,..,4$, we can compute the homography by solving the homogeneous equation \eqref{eq:dlt}. We compute the Singular Value Decomposition of A:
\begin{align}
    A &= U\Sigma V^T = U [\text{diag}(\sigma_1, \dots, \sigma_8)|0] V^T \\
      &= U \begin{pmatrix}
          \sigma_1  &  &\cdots & & 0 \\
           & \sigma_2 & \cdots & & 0 \\
            &  &\ddots &  & 0 \\
            &  &    & \sigma_8 & 0
      \end{pmatrix}
      \begin{pmatrix}
          v_1^T \\
          v_2^T \\
          \vdots \\
          v_8^T \\
          v_9^T
      \end{pmatrix} 
\end{align}
Since $A\in \mathbb{R}^{8\times9}$ it has $\text{rank}(A)=8$. We assume sorted singular values: $\sigma_1> \sigma_2> \dots >\sigma_8$. The solution to \eqref{eq:dlt} is then found in the corresponding row $v_9$ of $V^T$.
\begin{align}
    Av_9 &= U \begin{pmatrix}
          \sigma_1  &  &\cdots & & 0 \\
           & \sigma_2 & \cdots & & 0 \\
            &  &\ddots &  & 0 \\
            &  &    & \sigma_8 & 0
      \end{pmatrix} V^T v_9 \\
    &= U \begin{pmatrix}
          \sigma_1  &  &\cdots & & 0 \\
           & \sigma_2 & \cdots & & 0 \\
            &  &\ddots &  & 0 \\
            &  &    & \sigma_8 & 0
      \end{pmatrix} \begin{pmatrix}
          0 \\
          \vdots \\
          0 \\
          1
      \end{pmatrix} \\
      &=  U \cdot 0 = 0
\end{align}

In summary, the homography estimation is described by the pseudocode \ref{alg:dlt}.
\begin{algorithm}
\caption{Direct Linear Transform}\label{alg:dlt}
\begin{algorithmic}
\Require $P_i,P_j, X_1, \cdots, X_4$
\Ensure $H$
\For{$k \in \{1,2,3,4\}$}
\State $\mathbf{x}_k \gets P_iX_k = (x_k,y_k,w_k)$ 
\State $\mathbf{x}_k' \gets P_jX_k = (x_k',y_k',w_k')$ 
\State $A_k \gets \begin{pmatrix}
            0 & -w_k'\mathbf{x}_k^T & y_k'\mathbf{x}_k^T \\
            -w_k'\mathbf{x}_kT & 0 & x_k'\mathbf{x}_k^T
        \end{pmatrix}$
\EndFor
\State $A \gets \begin{pmatrix}
            A_1 \\
            \vdots\\
            A_4
        \end{pmatrix}$
\State $U,\Sigma, V^T \gets \mathbf{SVD}(A)$
\State $v_1, \cdots, v_9 \gets V$
\State $H_{ij} \gets \textbf{reshape}(v_9)$
\State \textbf{return} $H_{ij}$
\end{algorithmic}
\end{algorithm}

\subsection{Illumination}
\begin{figure}[h]
    \centering
    \includegraphics{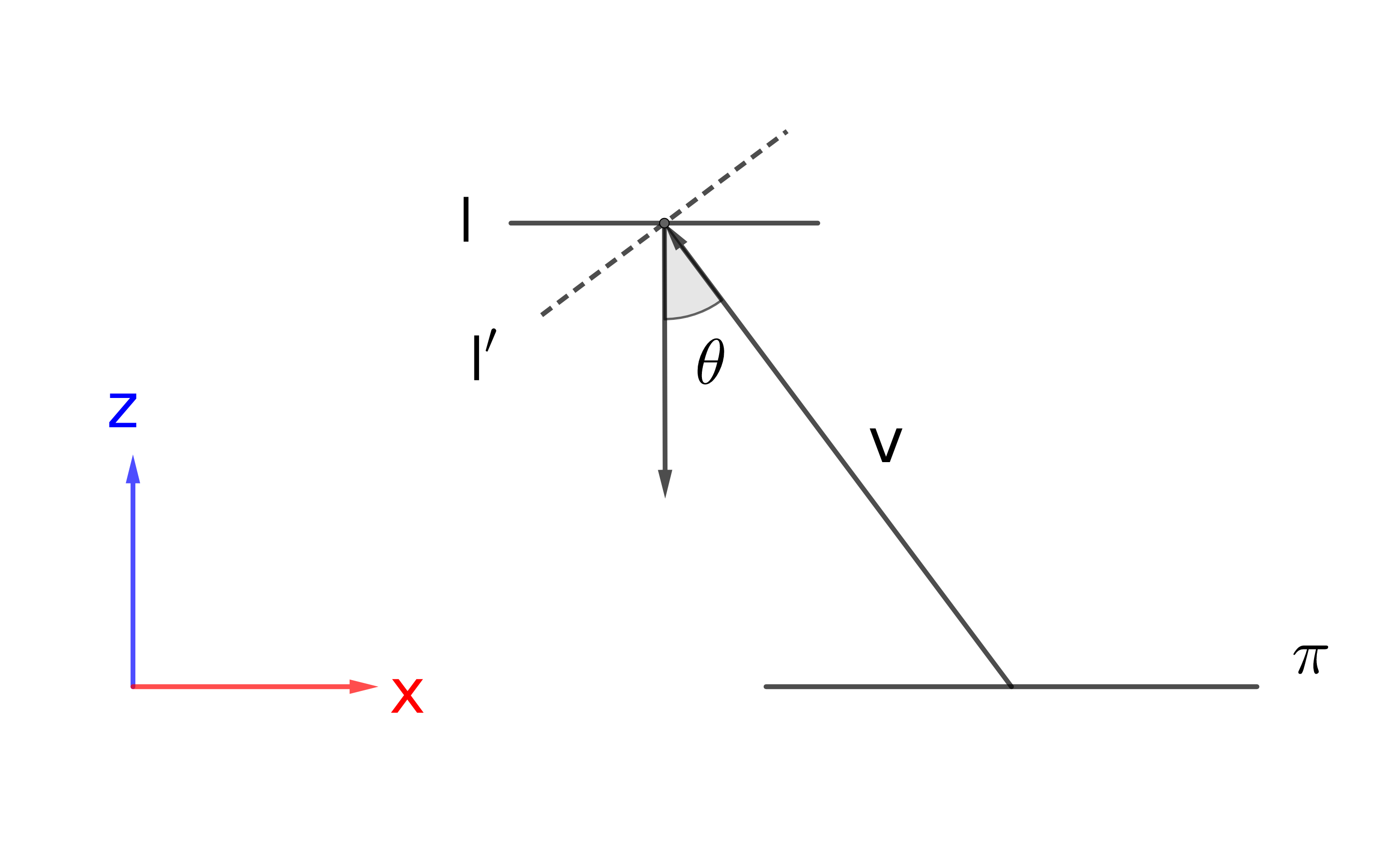}
    \caption{An illustration for the centering of an area light $l$. By default, Blender aligns new light sources along the $z$-axis. The correct orientation is described by the dotted line $l'$. $\theta$ describes the rotation offset between both configurations. $\theta$ is the angle between the (inverse) location vector  $-v$  of the light and the vector $(0,0,-1)$}
    \label{fig:light-orient}
\end{figure}

We add varying illumination by randomizing the light sources in the scene. The randomization can be described by the pseudocode \ref{alg:illumination}.
The light is randomly sampled from Blender's predefined light sources: spotlight, point light, and area light. We do not use a light source that simulates the sun because we want to simulate indoor environments. We randomize the intensity and color of the light. This randomization can create a large variety in the appearance of shadows, specularities, and the overall color scheme of the image. \\
The color is sampled using the HSV color space. Let $c=(H,S,V)$ be the color of the light. We sample $c$ using $H \sim \mathcal{U}[0,1],S \sim \mathcal{U}[0,\epsilon], V=1$.
$V$ is fixed because the intensity of the color is already affected by the light source itself. The hue is completely randomized to create random colors. However, the range of the saturation is limited to a small $\epsilon$ to generate lights that are closer to white. Having too high saturation creates unrealistic and extreme lighting conditions that make the reconstruction of the painting's color very ambiguous, while low saturation still introduces enough variance in the appearance of the painting. \\
Furthermore, we set the orientation of the light such that its direction is centered on the images. By default, Blender sets the orientation of the lights along the $z$-axis. As can be seen in figure \ref{fig:light-orient}, the correct orientation $\theta$ can be computed from the angle between the location vector $v$ and the z-axis $(0,0,1)$.  

\begin{algorithm}[h]
\caption{Add Illumination} \label{alg:illumination}
\begin{algorithmic}
\Require $n$
\Ensure $\{l_1,\cdots, l_n\}$
\State lights $\gets \emptyset$
\For{$k \in \{1,\cdots,n\}$}
\State type $\sim \mathbf{sample}\{\text{"Spot","Point","Area"}\}$ 
\State $l_k \gets$ new Light(type) \Comment{create light object} \\
\State $l_k$.intensity $\sim \mathcal{U}(I_{min},I_{max})$ \Comment{random intensity} \\
\State $x,y,z \sim \mathcal{U}(x_{min},x_{max})\times \mathcal{U}(y_{min},y_{max})\times \mathcal{U}(z_{min},z_{max})$  \Comment{random position} 
\State $l_k$.position $\gets x,y,z$ \\
\State h,s,v $\gets \mathcal{U}(0,1)\times \mathcal{U}(0,\epsilon)\times {1}$ \Comment{sample color in HSV space}
\State $l_k$.color $\gets HSV(h,s,v) $\\
\State $v \gets l_k.\text{position}$ \Comment{get location vector of light}
\State $l_k$.orientation $\gets \measuredangle(v, (0,0,1)^T)$  \Comment{center the direction by computing the angle between $v$ and z-axis.}\\
\State lights.append($l_k$)
\EndFor
\State \textbf{return} lights
\end{algorithmic}
\end{algorithm}

\subsection{Add Geometric Objects}

\begin{figure}[h]
    \centering
     \begin{subfigure}[b]{0.49\textwidth}
     \centering
     \includegraphics[width=\textwidth]{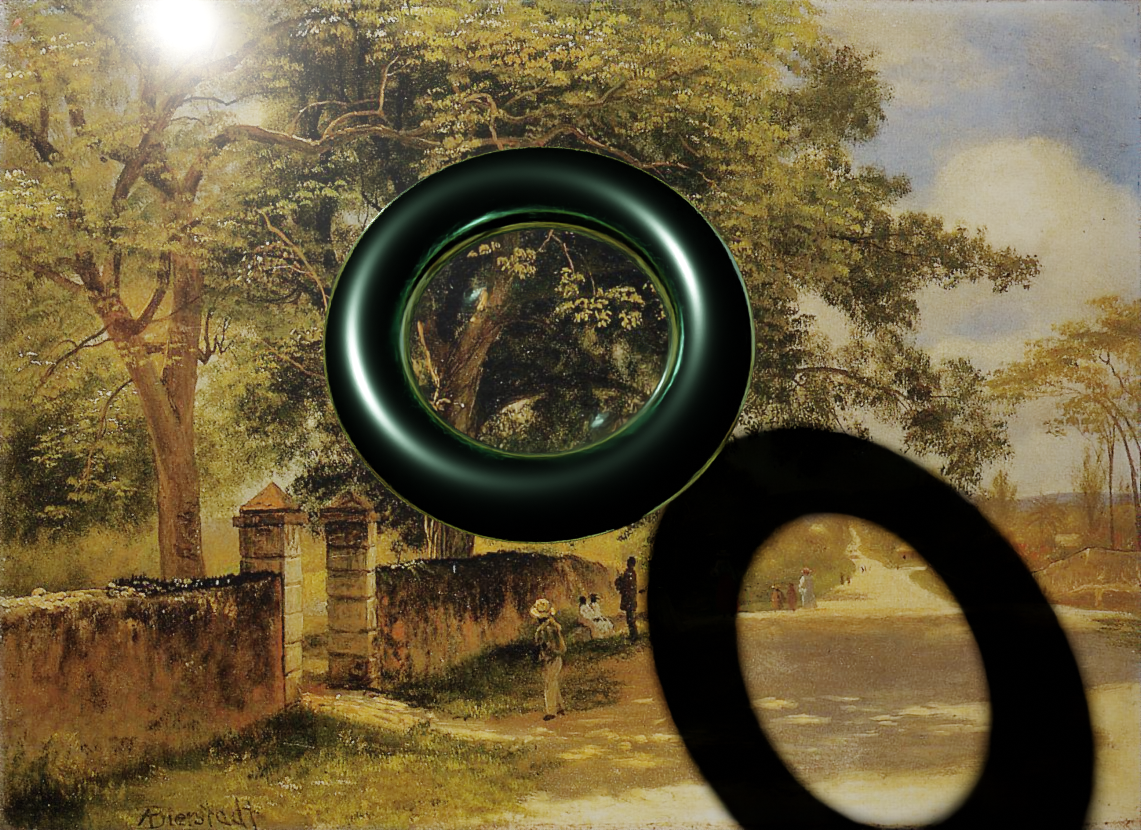}
     \caption{Gloss}
     \label{fig:gloss}
     \end{subfigure}
     \begin{subfigure}[b]{0.49\textwidth}
     \centering
     \includegraphics[width=\textwidth]{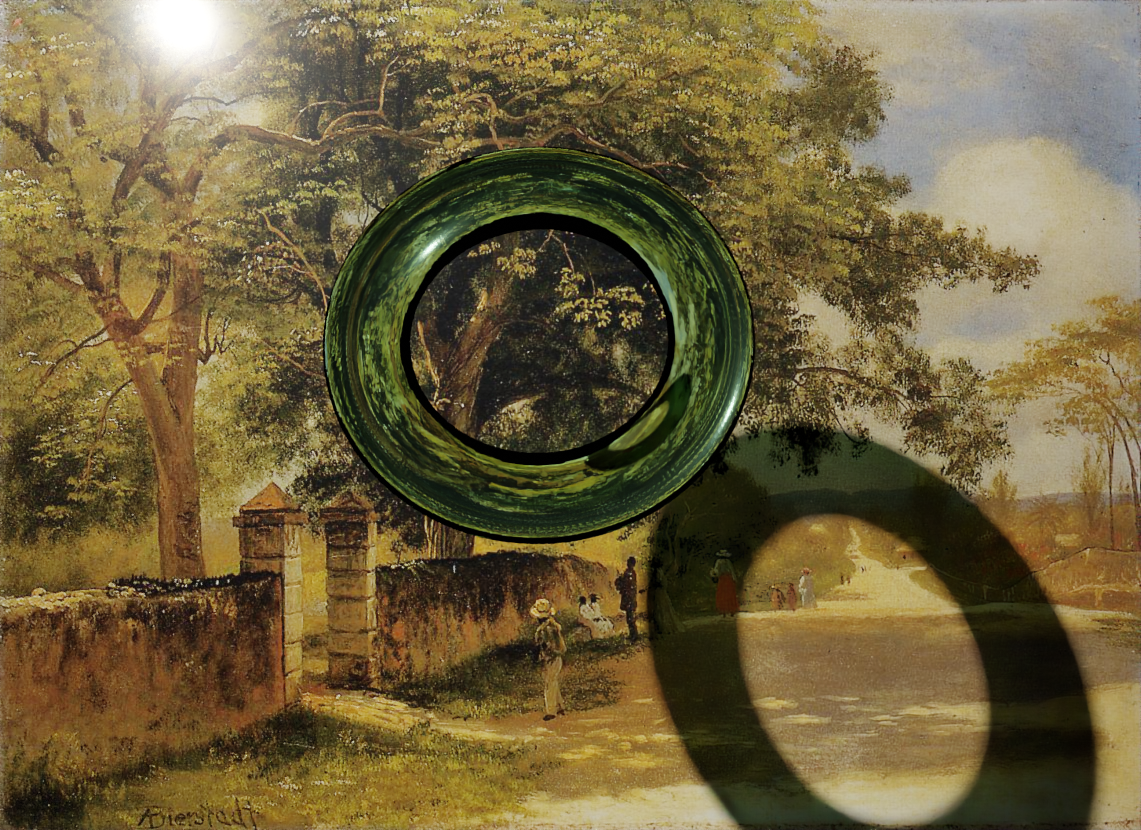}
     \caption{Refraction}
     \label{fig:refraction}
     \end{subfigure}
     \begin{subfigure}[b]{0.49\textwidth}
     \centering
     \includegraphics[width=\textwidth]{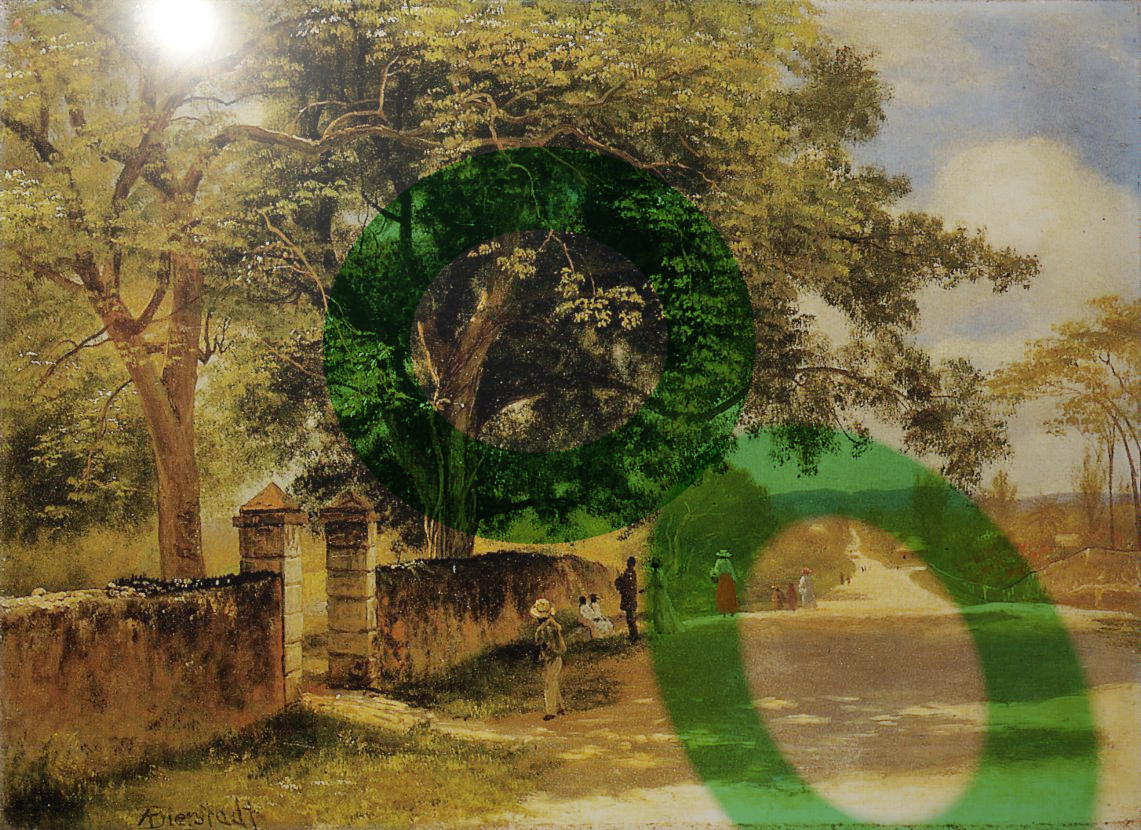}
     \caption{Transparency}
     \label{fig:transparent}
     \end{subfigure}
     \begin{subfigure}[b]{0.49\textwidth}
     \centering
     \includegraphics[width=\textwidth]{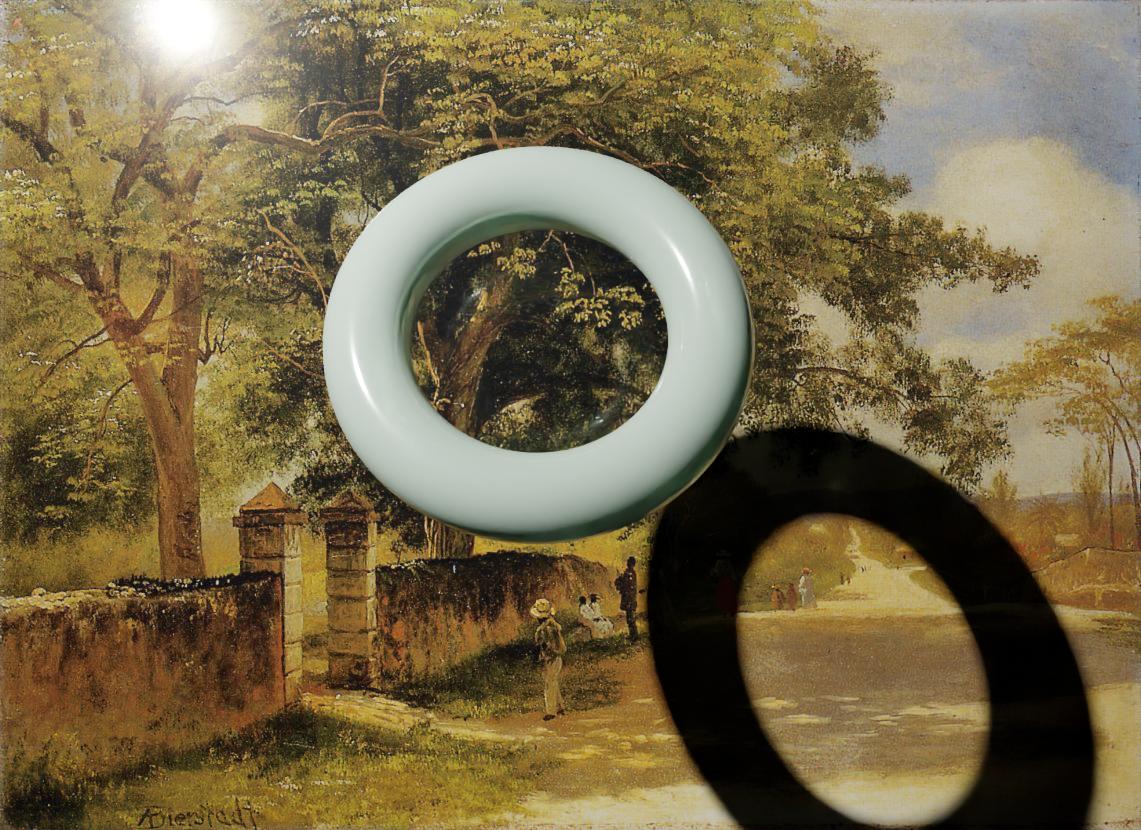}
     \caption{Solid Material}
      \label{fig:principledBSDF}
     \end{subfigure}
    \caption{A randomly generated torus under the same lighting conditions but with varying materials. }
    \label{fig:material}
\end{figure}

Randomized geometric objects are added in the space between light sources and the image plane. The purpose of these objects is to create occlusions and shadows. The objects obfuscate image content from the cameras and block light from reaching the image plane. The material of the object is also chosen from Blender's shaders. Depending on the material, light is either completely blocked, refracted, or color filtered. As can be seen in figure \ref{fig:material}, the material affects the appearance of the image content and casts a shadow. Materials in figure \ref{fig:gloss} and \ref{fig:principledBSDF} create hard shadows, while figures \ref{fig:refraction} and \ref{fig:transparent} create softer shadows and even change the shadow's color.
Transparent materials also do not fully obfuscate the image content.

\begin{algorithm}[H]
\caption{Add Geometric Objects} \label{alg:geom}
\begin{algorithmic}
\Require $n$
\Ensure $\{o_1,\cdots, o_n\}$
\State objects $\gets \emptyset$
\For{$k \in \{1,\cdots,n\}$}
\State geometry $\sim \mathbf{sample}\{\text{"Sphere","Cube","Cylinder",}\dots\}$ 
\State $o_k \gets$ new Object(type) \\
\State $x,y,z \sim \mathcal{U}(x_{min},x_{max})\times \mathcal{U}(y_{min},y_{max})\times \mathcal{U}(z_{min},z_{max})$  \Comment{random position}
\State $o_k$.positions $\gets (x,y,z)$ \\
\State $\alpha_x,\alpha_y,\alpha_z \sim \mathcal{U}(0,2\pi)\times \mathcal{U}(0,2\pi)\times \mathcal{U}(0,2\pi)$  \Comment{random orientation}
\State $o_k$.rotation $\gets (\alpha_x,\alpha_y,\alpha_z)$ \\
\State $s_x,s_y,s_z \sim \mathcal{U}(s_{x_{min}},s_x{_{max}})\times \mathcal{U}(s_y{_{min}},s_y{_{max}})\times \mathcal{U}(s_z{_{min}},s_z{_{max}})$  \Comment{random scaling}
\State $o_k$.scale $\gets(s_x,s_y,s_z)$ \\

\State material $\sim \mathbf{sample}\{\text{"Refraction","Transparent","Metallic",}\dots\}$ \Comment{sample random material type}
\State $o_k$.material $\gets $ new Material(material) \\
\State objects.append($o_k$)
\EndFor
\State \textbf{return} objects
\end{algorithmic}
\end{algorithm}

\subsection{Rendering}
\label{sec:rendering}
\begin{figure}
    \centering
     \begin{subfigure}[b]{0.24\textwidth}
     \centering
     \caption{0.1s}
     \includegraphics[width=\textwidth]{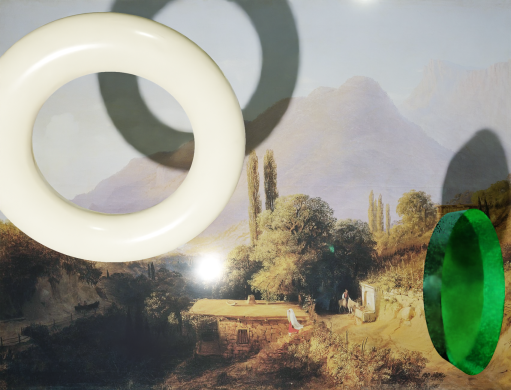}
     \label{fig:01s}
     \end{subfigure}
     \begin{subfigure}[b]{0.24\textwidth}
     \centering
      \caption{1s}
     \includegraphics[width=\textwidth]{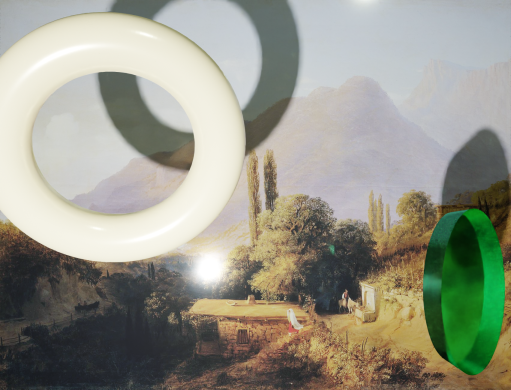}
     \label{fig:1s}
     \end{subfigure}
     \begin{subfigure}[b]{0.24\textwidth}
     \centering
     \caption{10s}
     \includegraphics[width=\textwidth]{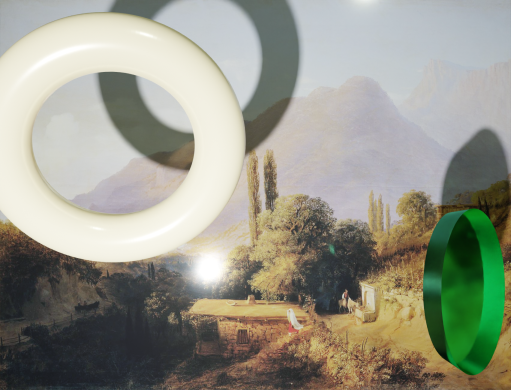}
     \label{fig:10s}
     \end{subfigure}
     \begin{subfigure}[b]{0.24\textwidth}
     \centering
  \caption{~5min}   
     \includegraphics[width=\textwidth]{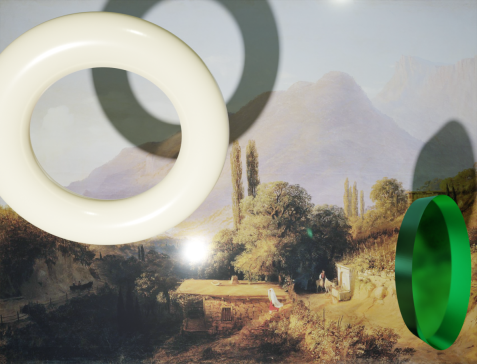}
      \label{fig:5m}
     \end{subfigure}
         \begin{subfigure}[b]{\textwidth}
     \centering
     \includegraphics[width=\textwidth]{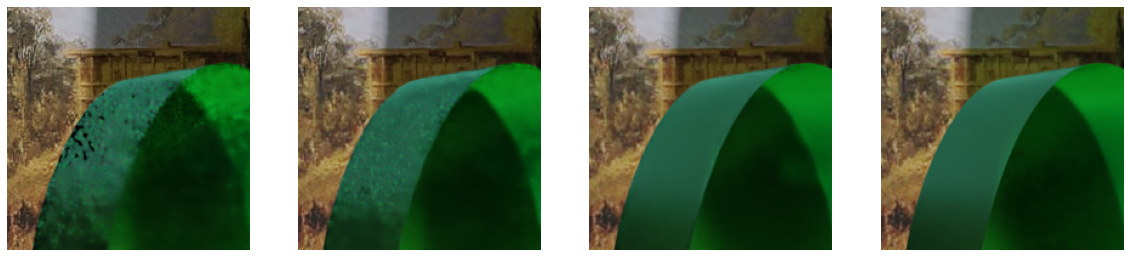}
     \includegraphics[width=\textwidth]{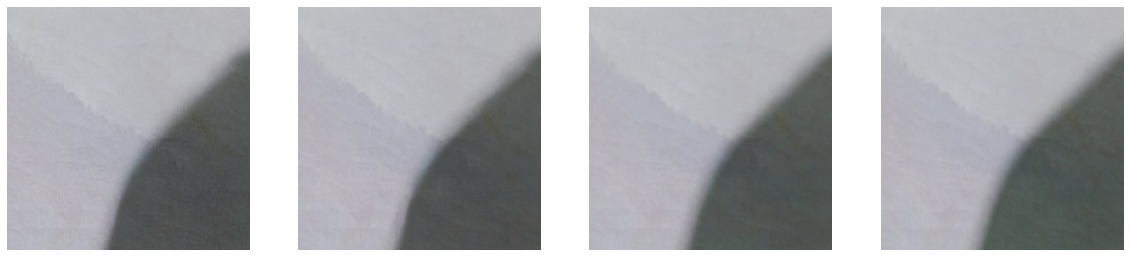}
     \includegraphics[width=\textwidth]{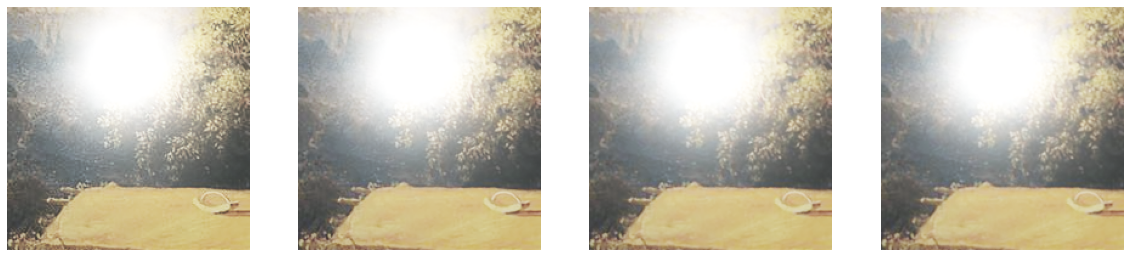}
      \label{fig:zoom}
     \end{subfigure}
      \caption{The images show the same scene rendered with different time limits. The first row shows the whole images, while the second to fourth rows show specific parts of the image.}
    \label{fig:qualityvstime}
\end{figure}

After a scene is configured with randomized lighting, occlusions, and cameras, an image is rendered using path tracing. Blender's physics-based path tracer is used to render the image from the perspective of a specific camera. Path tracing allows the creation of more realistic shadows and lighting effects compared to rasterization. Especially effects such as transparency, reflection, and refraction can be realistically modeled using path tracing by sampling light rays. %TODO: citation
Path tracing simulates the physical image formation process much closer than rasterization techniques. 
Rasterization requires the use of various techniques, such as texture maps and shadow maps, to approximate the same effects.%TODO: citation
The number of rays cast to estimate the light distribution of the scene and the resulting image can be limited using a time constraint.
A time limit is set to balance the image quality and the amount of data that can be efficiently generated. Figure \ref{fig:qualityvstime} illustrates the effect of rendering time on the resulting image. The results do not differ too much. Shadows, illumination, and occlusions are visualized correctly, even when using fewer rays. A low amount of sampling can create aliasing effects, as can be seen in the glassy material. In comparison, more sampling creates more realistic effects. 

\begin{figure}
    \centering
    \begin{subfigure}[b]{0.32\textwidth}
         \centering
         \includegraphics[width=\textwidth]{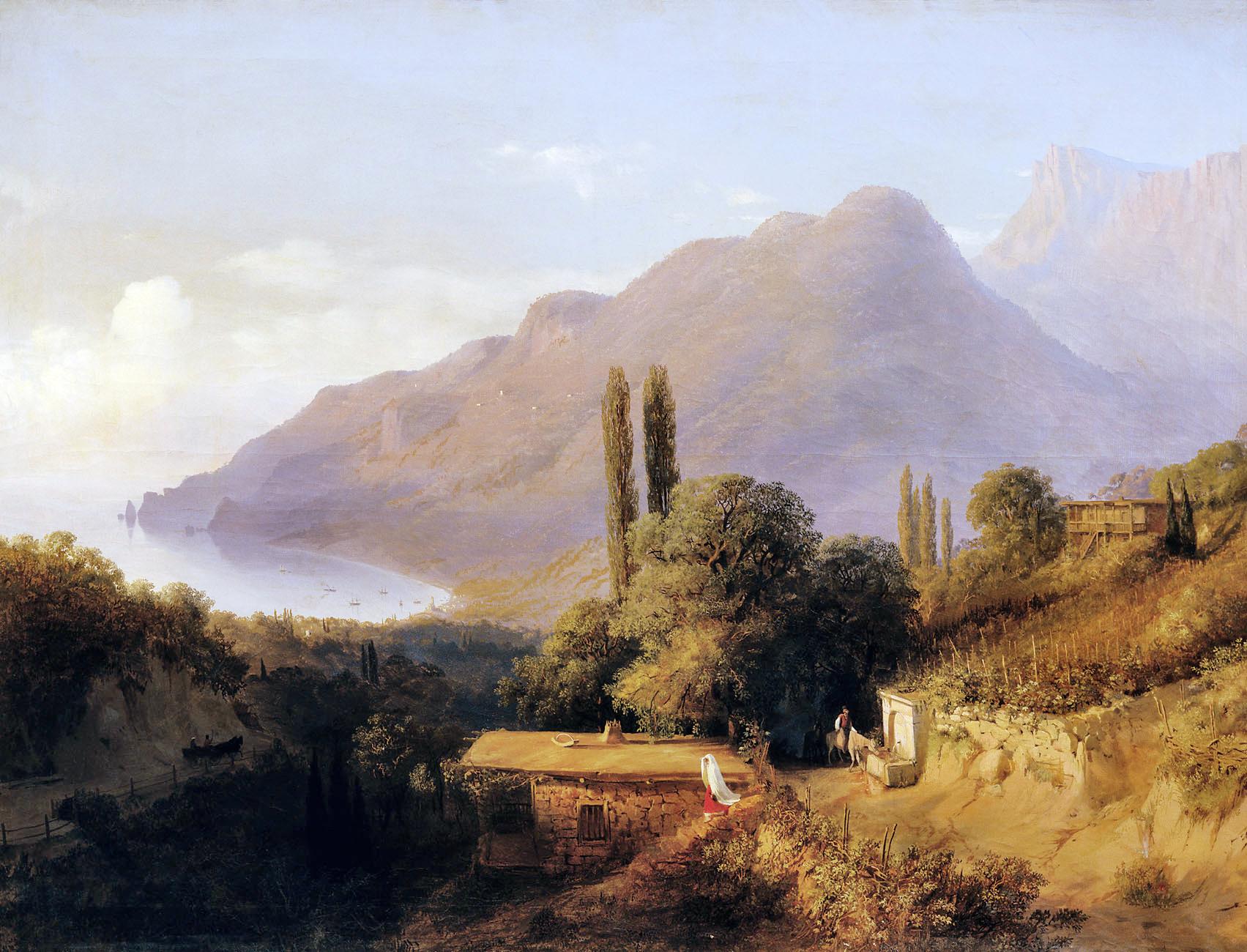}
            \caption{Original image: $I_O$}
         \label{fig:orig}
         \end{subfigure}
    \begin{subfigure}[b]{0.32\textwidth}
         \centering
         \includegraphics[width=\textwidth]{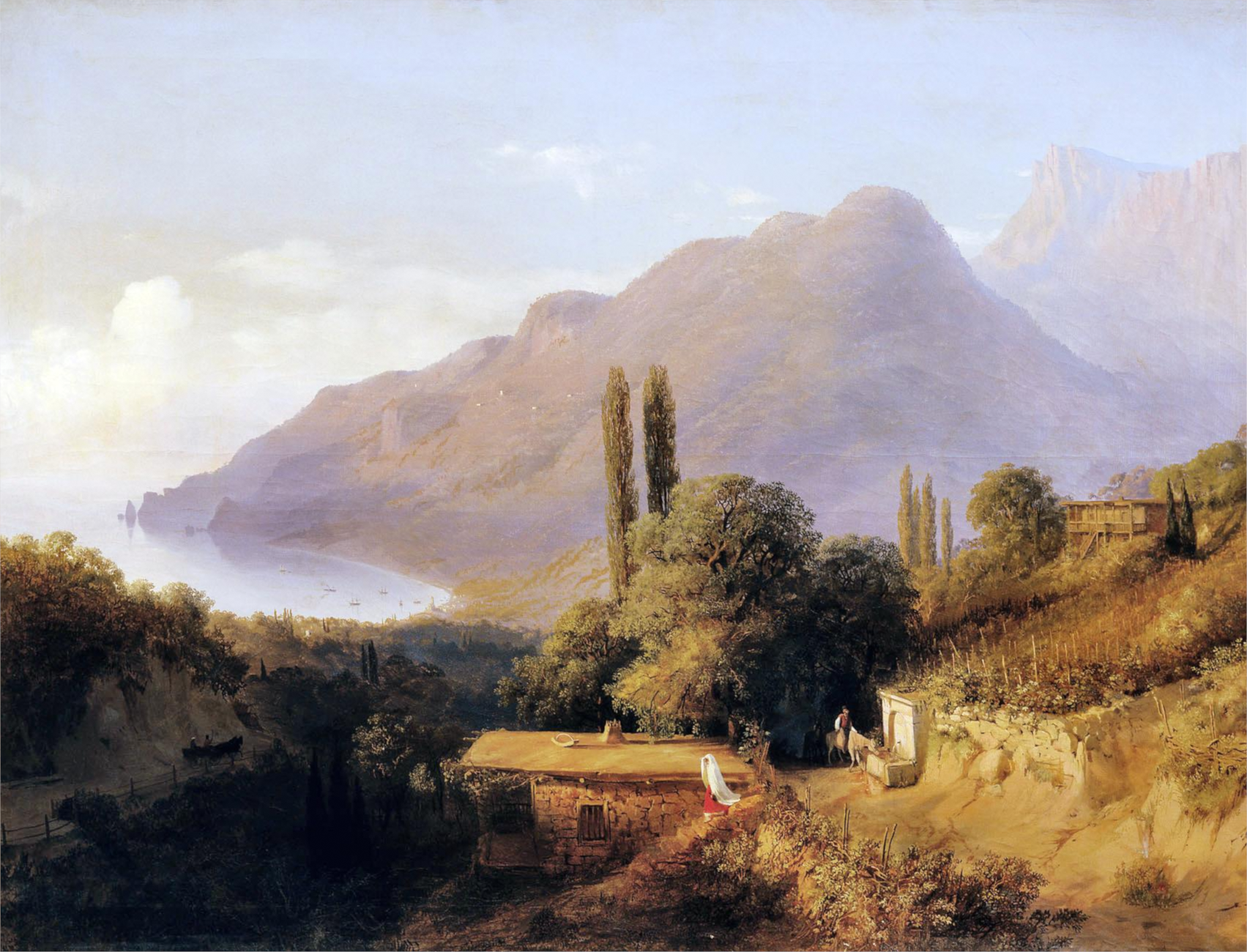}
          \caption{Ambient lighting: $I_A$}
         \label{fig:renderGT}
     \end{subfigure}
    \begin{subfigure}[b]{0.32\textwidth}
         \centering
         \includegraphics[width=\textwidth]{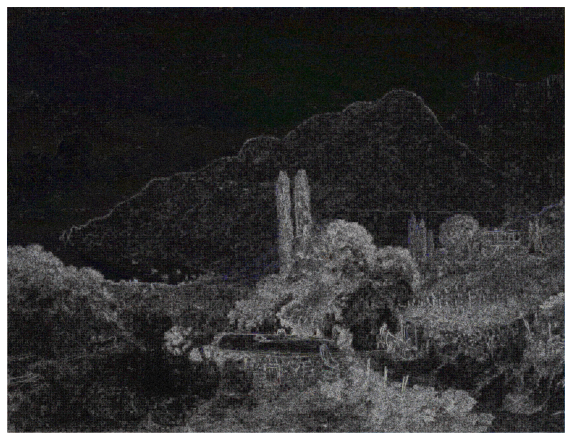}
            \caption{Difference: $|I_O-I_A|$}
         \label{fig:diff}
         \end{subfigure}
    \caption{The images show the original image and the output image generated by the rendering pipeline using an aligned camera and ambient lighting. c) shows the absolute difference of each pixel.}
    \label{fig:label}
\end{figure}

Each rendered image describes a distorted data point of the original image. Path tracing is also used to generate the ground truth label. It is possible to use the original image as a ground truth label. However, the rendering pipeline can create bias and small misalignment. For this reason, the label is created under similar conditions as the distorted images using ambient illumination without occlusion. A virtual camera is used that is aligned with the image as described in chapter \ref{chap:alignment}. The resulting image is free of artifacts. Figure \ref{fig:label} compares the original images and the rendered image. Although both images look very similar, there is a small difference between both images around edges, as can be seen in figure \ref{fig:diff}. The difference might be caused by small misalignments or texturing artifacts.

\begin{figure}
    \centering
    \includegraphics[width=0.49\textwidth]{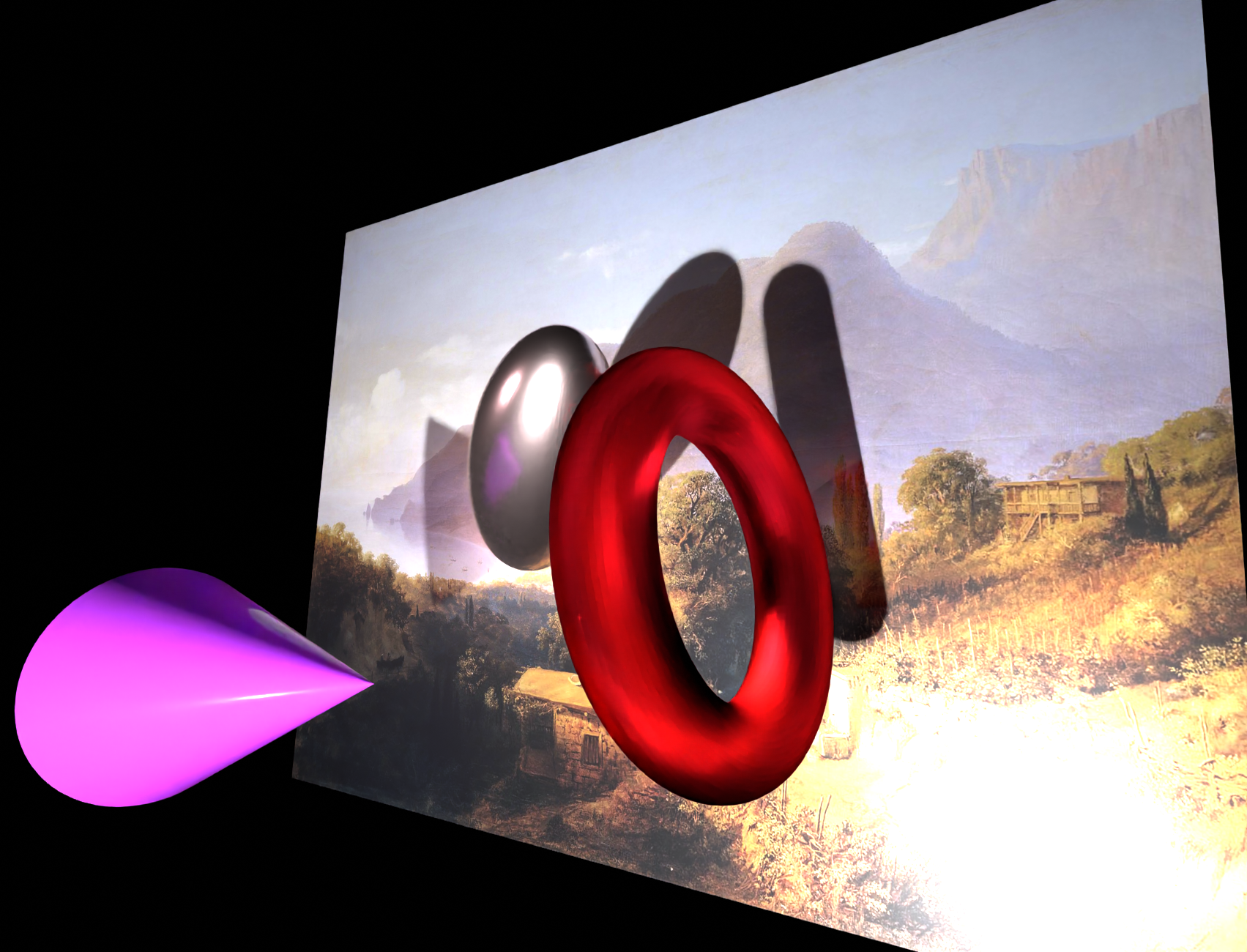}
    \includegraphics[width=0.49\textwidth]{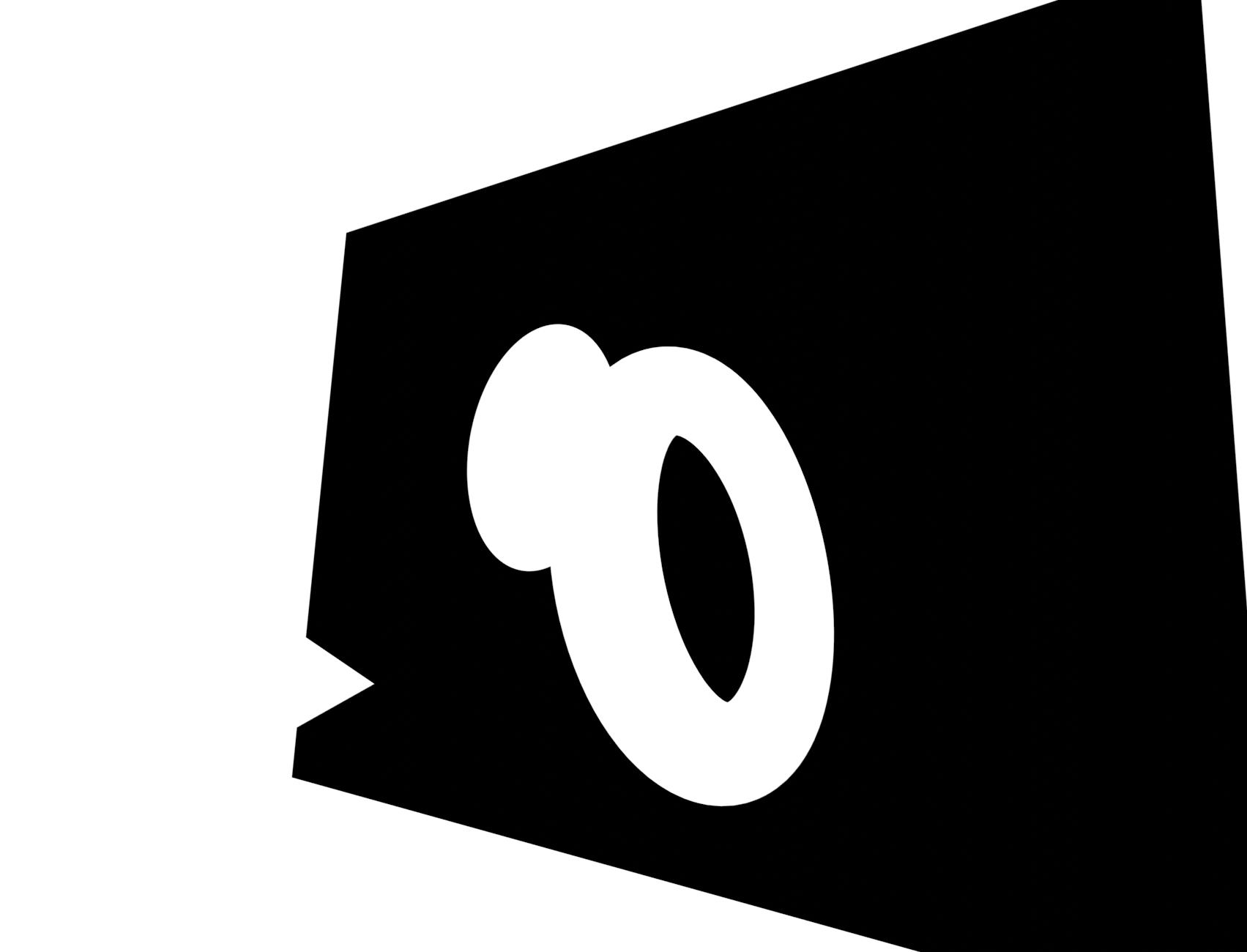}
    \caption{The left image shows a randomized scene. The right image shows the same scene rendered after changing the plane to a diffuse black and all geometric objects to a diffuse white color.}
    \label{fig:mask}
\end{figure}

In order to differentiate between distorted parts of the image (caused by lighting, shadows, and specular highlights) and occlusions, a segmentation mask is computed that separates foreground and background pixels. The mask should be aligned with the corresponding camera. After rendering any scene, a corresponding occlusion mask is rendered using the same camera and geometric objects. For any given scene, the material of the plane is changed to a diffuse black. All objects are changed to a diffuse white material. We also use ambient illumination. The segmentation mask is rendered using regular rasterization.  Figure \ref{fig:mask} shows a mask generated from a given scene. The resulting image is binary, with black pixels describing parts of the painting and white pixels describing occlusions or background.

\section{Datasets}

Using the rendering pipeline from chapter \ref{chap:render}, we create two datasets. By aligning the camera with the image, we create a dataset of aligned sequences without perspective distortion. Another dataset is created by adding perspective distortions. 

\subsection{Aligned Dataset}
\label{sec:aligneddata}
\begin{figure}[h]
    \centering  
    \includegraphics[width=0.19\textwidth]{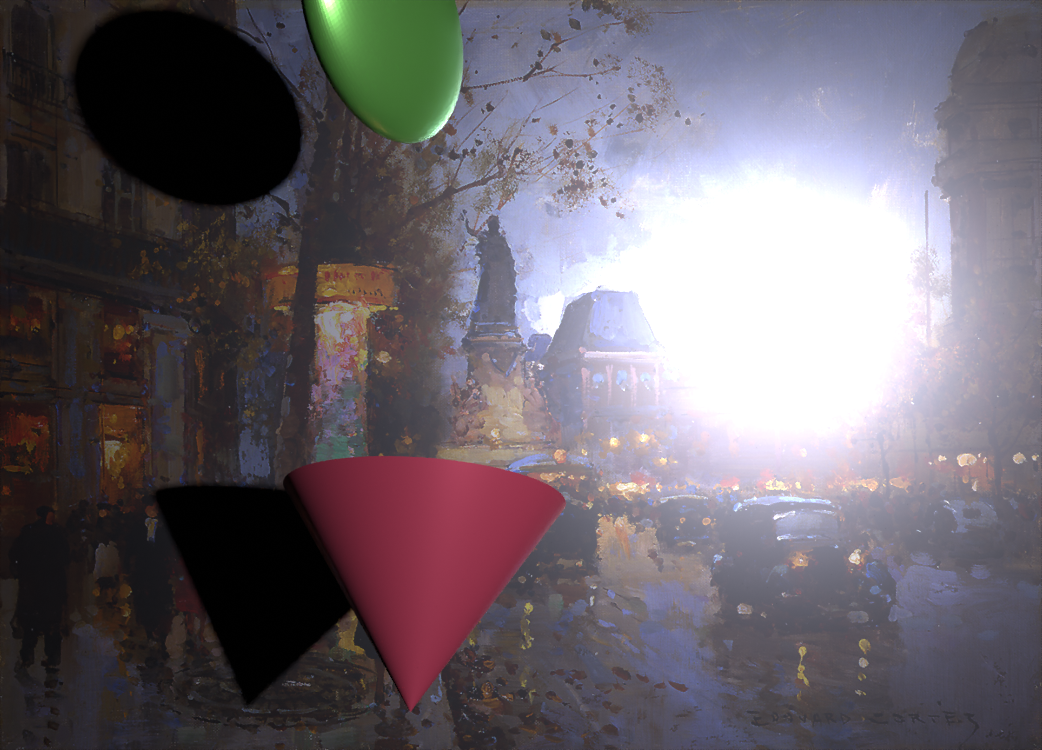}
    \includegraphics[width=0.19\textwidth]{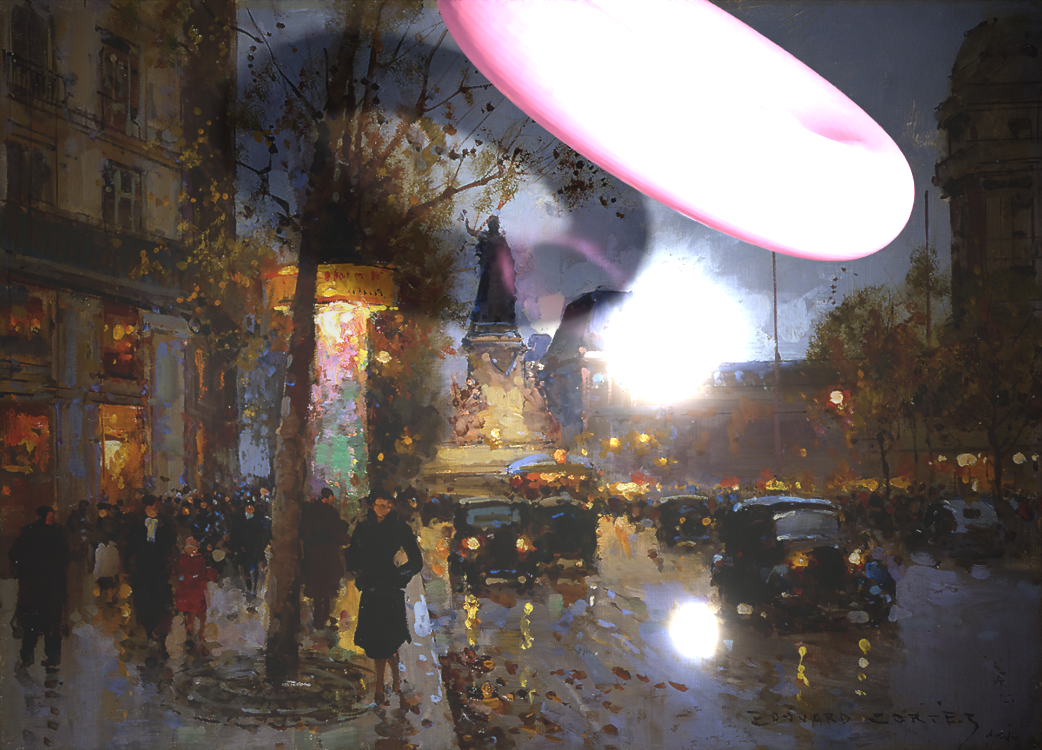}
    \includegraphics[width=0.19\textwidth]{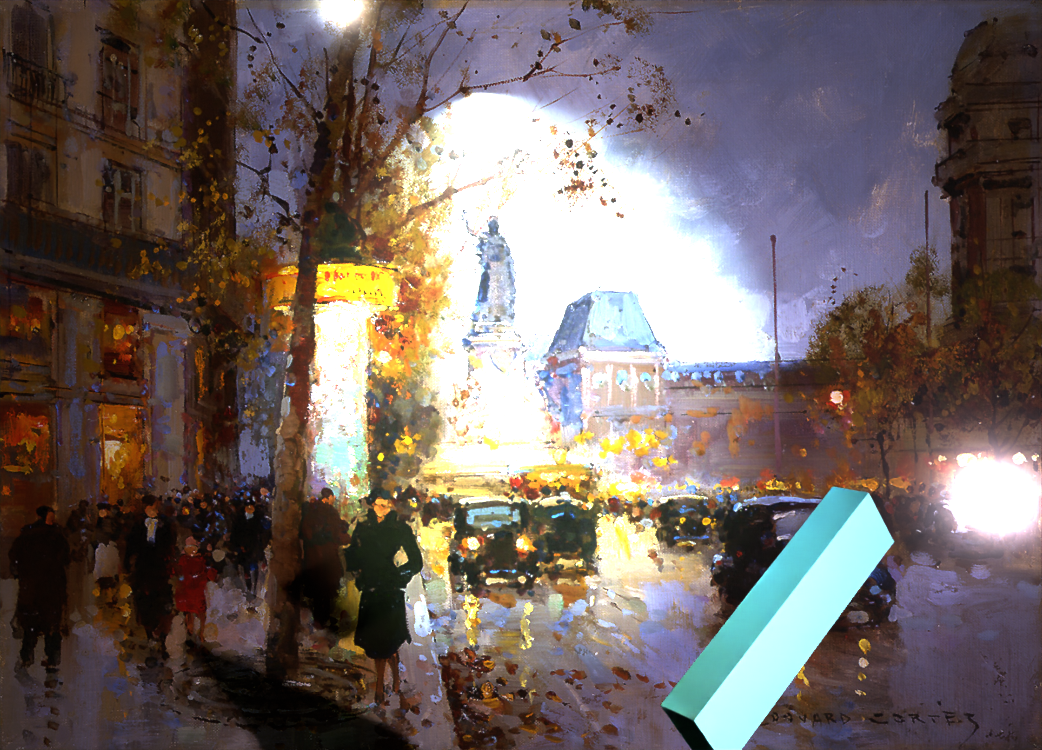}
    \includegraphics[width=0.19\textwidth]{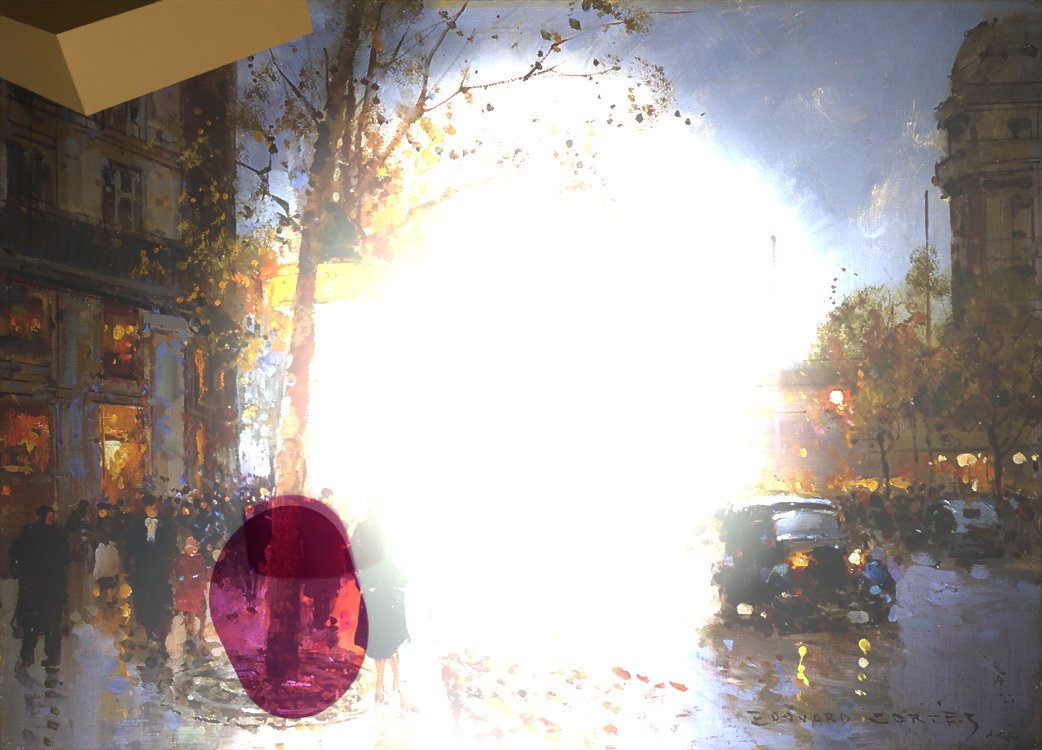}
    \includegraphics[width=0.19\textwidth]{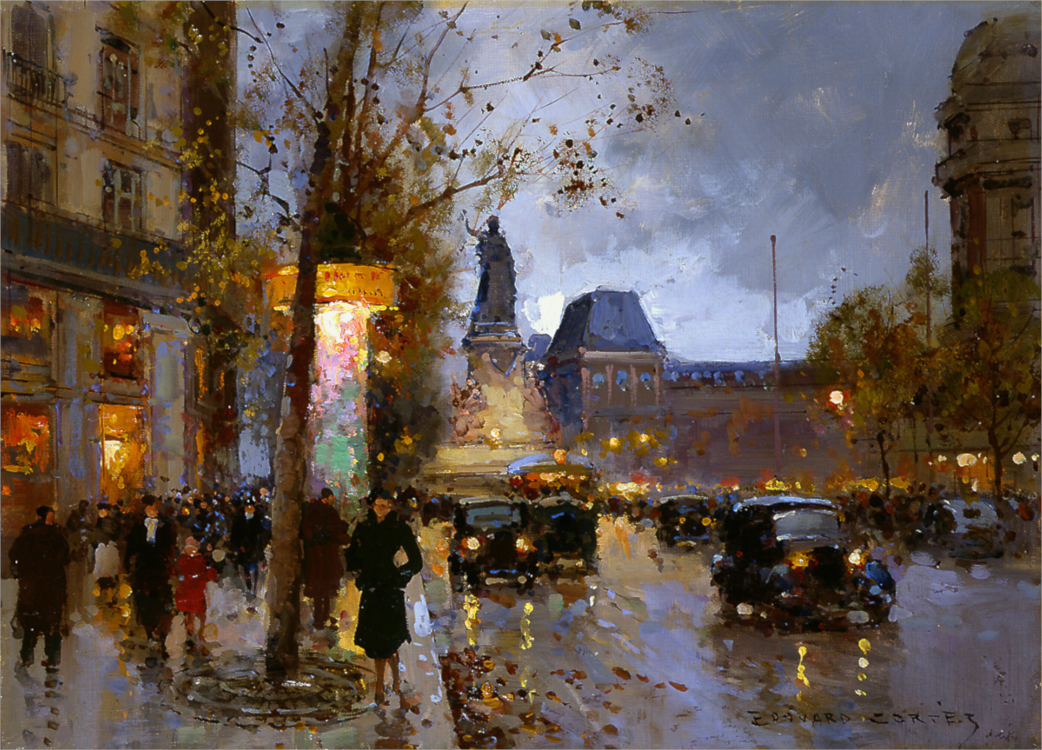} 
    
    \includegraphics[width=0.19\textwidth]{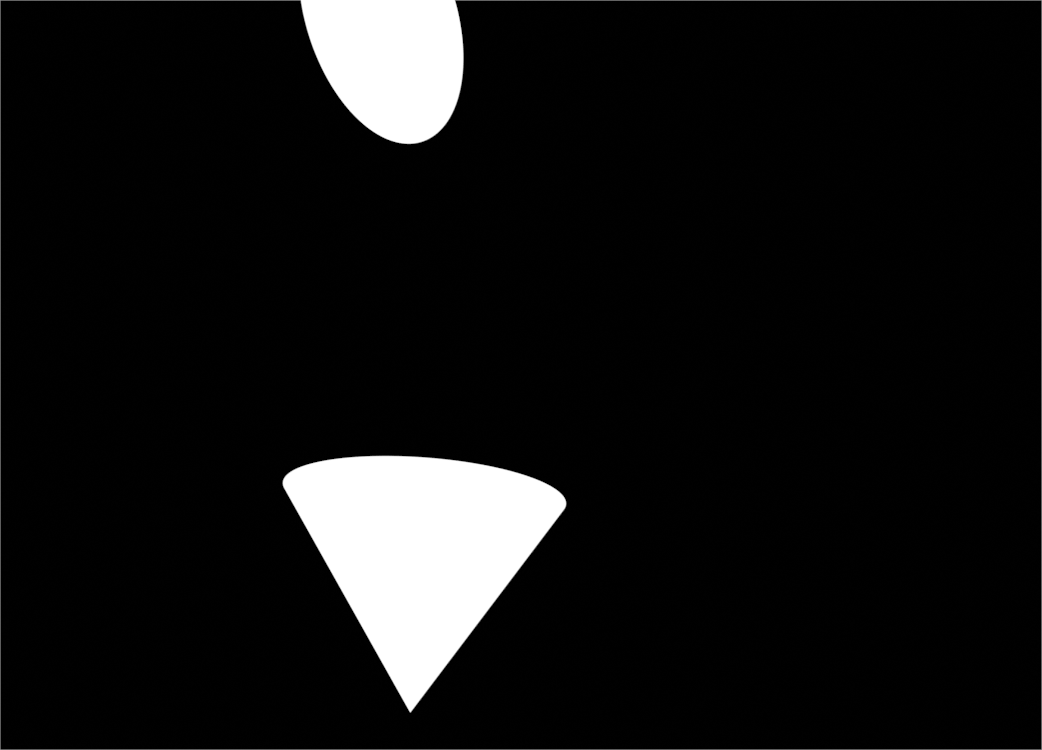}
    \includegraphics[width=0.19\textwidth]{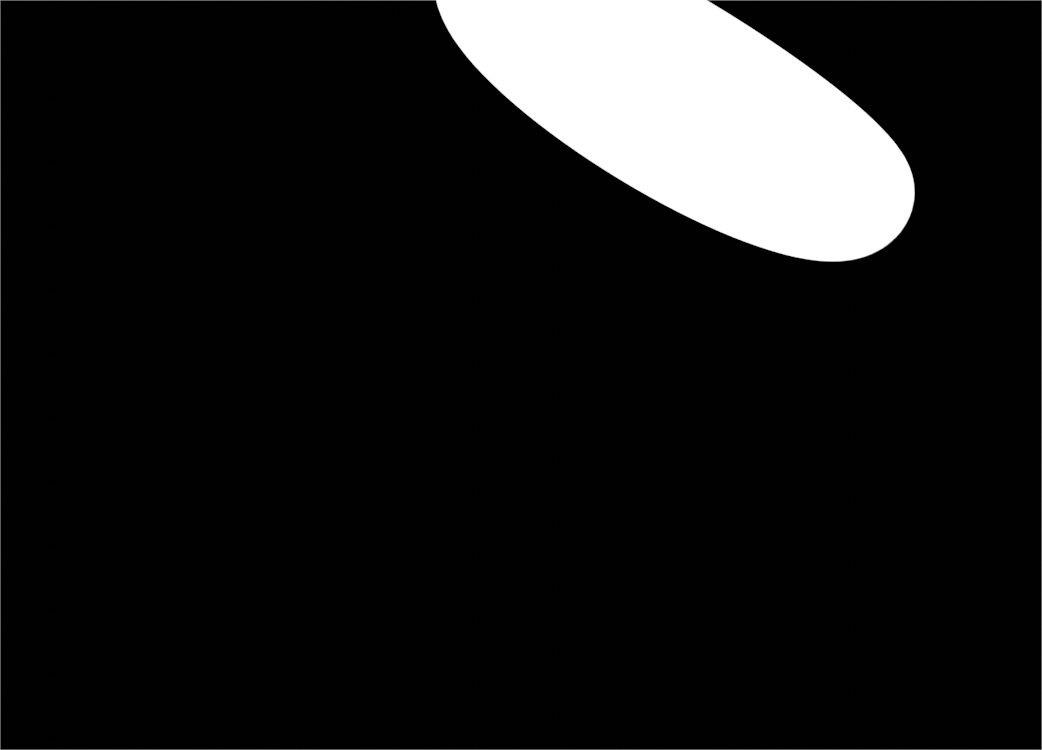}
    \includegraphics[width=0.19\textwidth]{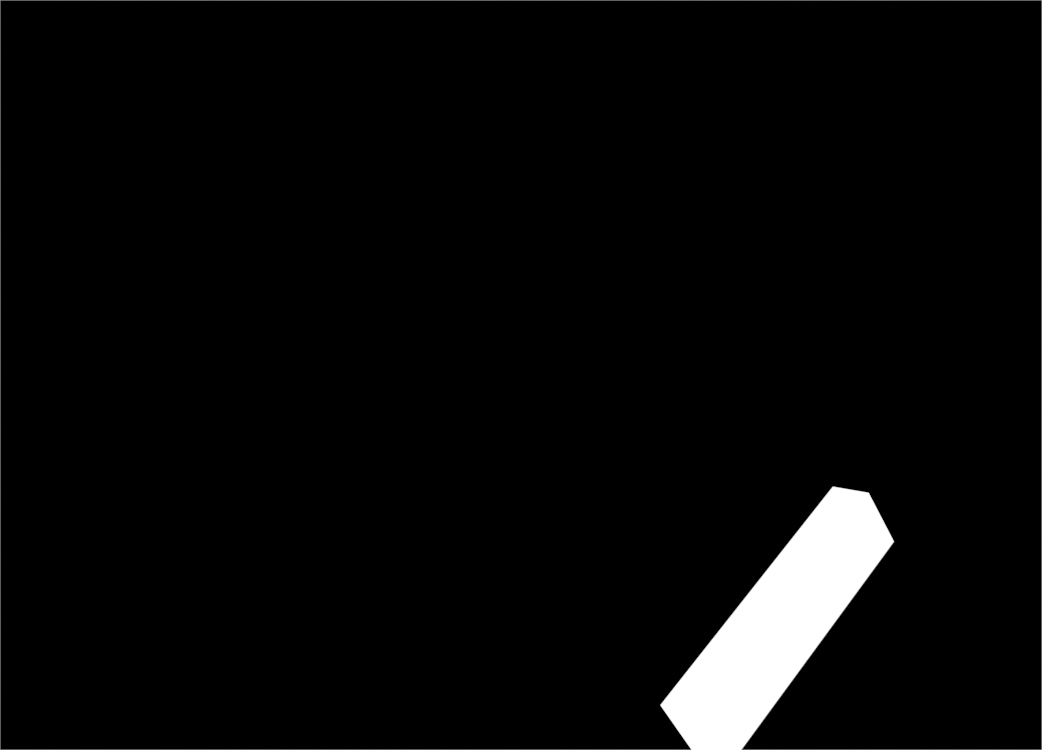}
    \includegraphics[width=0.19\textwidth]{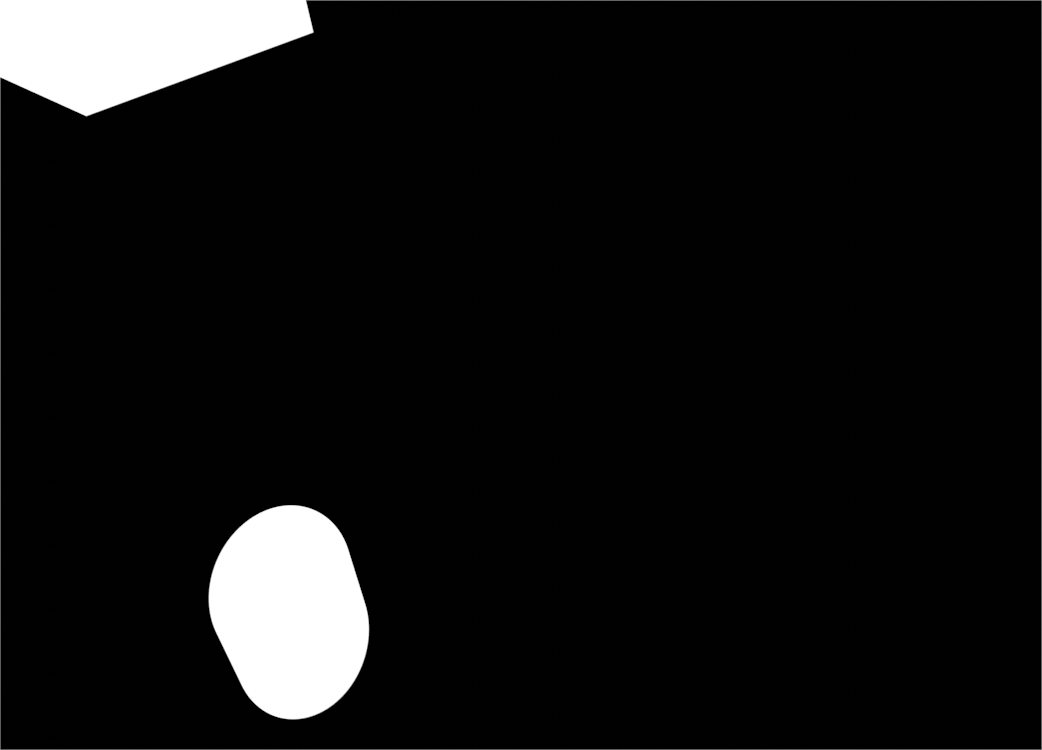}
    \includegraphics[width=0.19\textwidth]{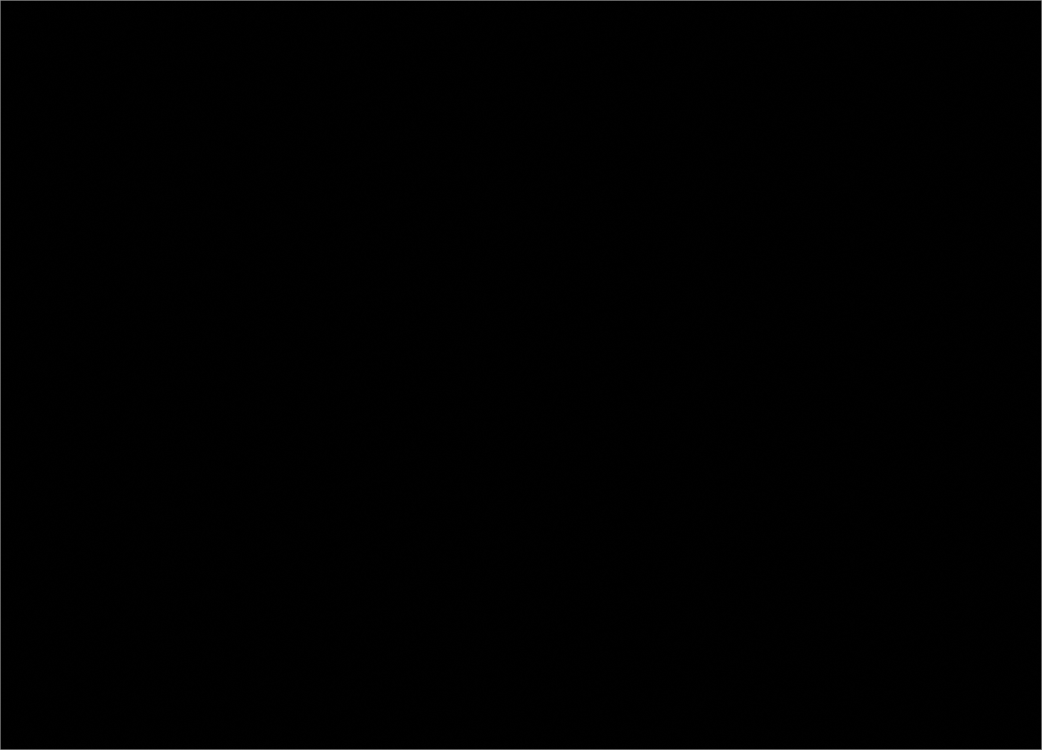}
    \caption{The top row shows four generated images without perspective distortions and with the corresponding label. The bottom row shows the occlusion mask for each image.}
    \label{fig:aligned}
\end{figure}

Figure \ref{fig:aligned} shows an example sequence of distorted images with the corresponding ground truth label. We generate image sequences, each containing multiple distorted images as our dataset. For each distorted image, an additional occlusion mask is created. The dataset can be used for multiple tasks. Since the amount of data can be increased indefinitely, it is especially useful for deep learning models.  

\subsubsection*{Image Restoration}
Using the aligned dataset, a model can be trained to remove artifacts from images. This can be done by a sequential model that learns to aggregate information from multiple images, such as Deep Sets \cite{zaheer2017deep,kwiatkowski2022specularity}
Alternatively, a model can be trained to remove artifacts from single images, e.g., single-image shadow removal \cite{qu2017deshadownet}. %TODO: Citations
Existing methods often deal with each type of artifact individually, whereas this dataset allows combining multiple artifacts simultaneously. The data generation can be configured to generate specific artifacts or combinations of them. Using the occlusion mask, inpainting methods can be trained to reconstruct the content of occluded regions. 

\subsubsection*{Background Subtraction}
The dataset can be used for background subtraction tasks. The artifact-free image describes the underlying background, while lighting and occlusion create distortions.  A model can be trained to detect occlusions and variations from the underlying background model. The generated occlusion masks can be used for learning foreground-background segmentation. Existing datasets consist of videos with a few scenes, such as CDNet \cite{goyette2012changedetection}. Although the videos provide a lot of data, there is little variation within each scene. Our dataset generation allows the creation of a large variety of different scenes and also increases the variation within each scene. This is useful to enable models to generalize over different backgrounds and make the background modeling more robust to changing lighting conditions.

\subsubsection*{Representation Learning}
Data augmentations, such as noise, blurring, random cropping, and geometric transformations, are used to create more robust representation learning or self-supervised training \cite{chen2020simple,bansal2022cold}. Using our dataset, a representation can be learned that is invariant to illumination and occlusions. Alternatively, a representation can be learned that disentangles the image content from lighting, shadows, occlusions, etc. 

\subsection{Misaligned Dataset}
\label{sec:misaligned}
\begin{figure}[h]
    \centering  
    \includegraphics[width=0.19\textwidth]{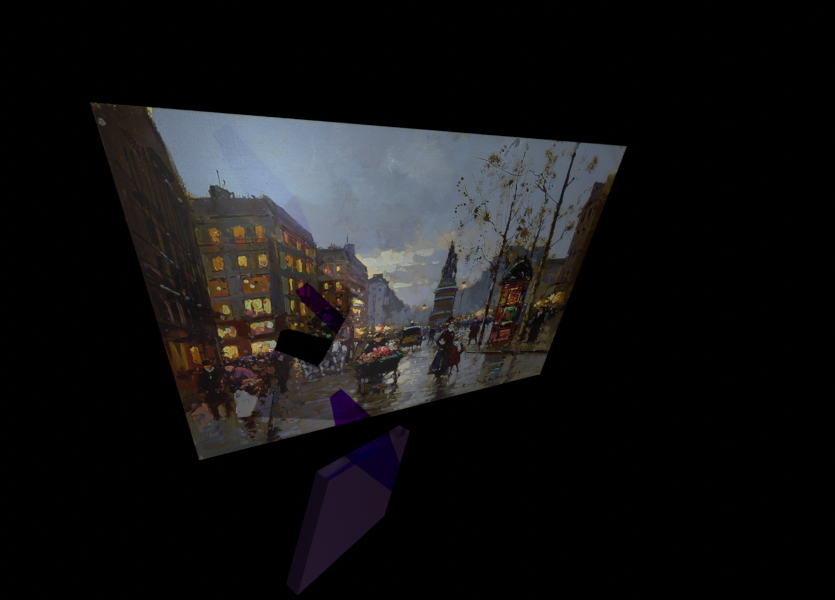}
    \includegraphics[width=0.19\textwidth]{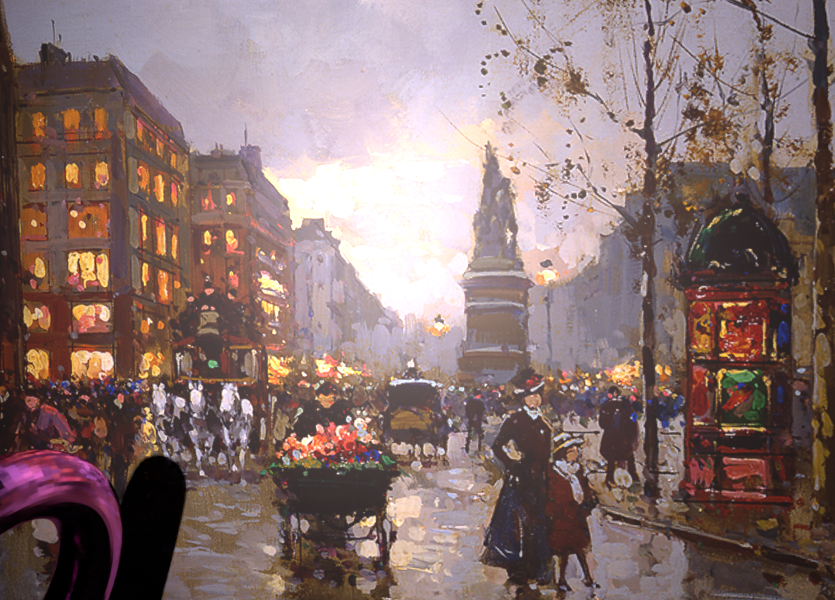}
    \includegraphics[width=0.19\textwidth]{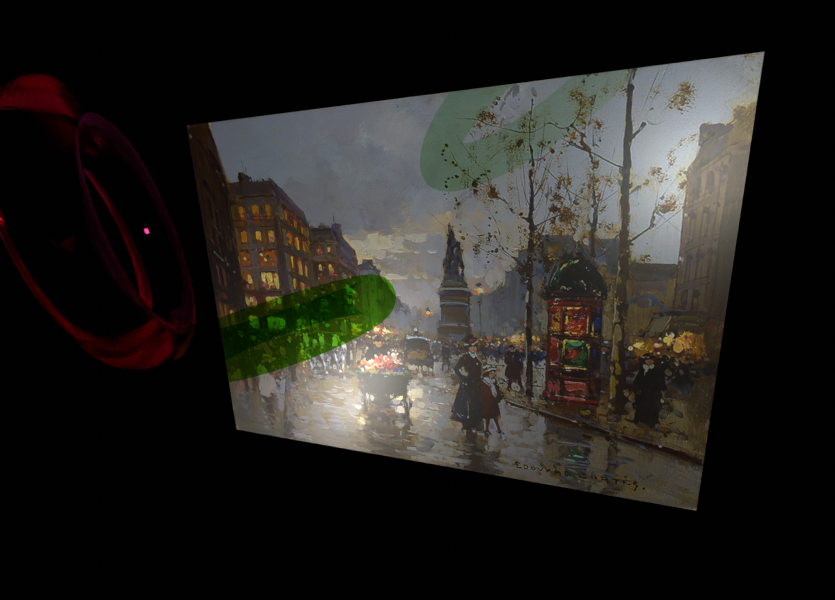}
    \includegraphics[width=0.19\textwidth]{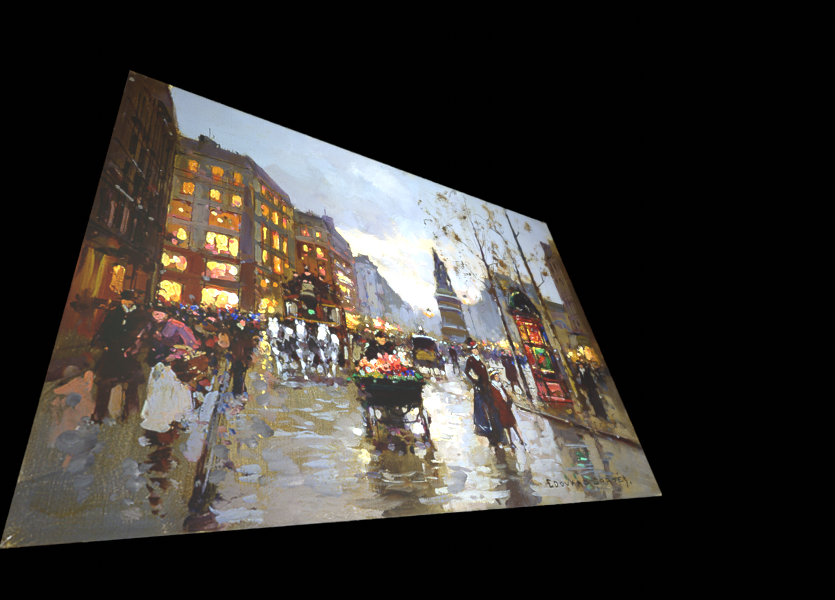}
    \includegraphics[width=0.19\textwidth]{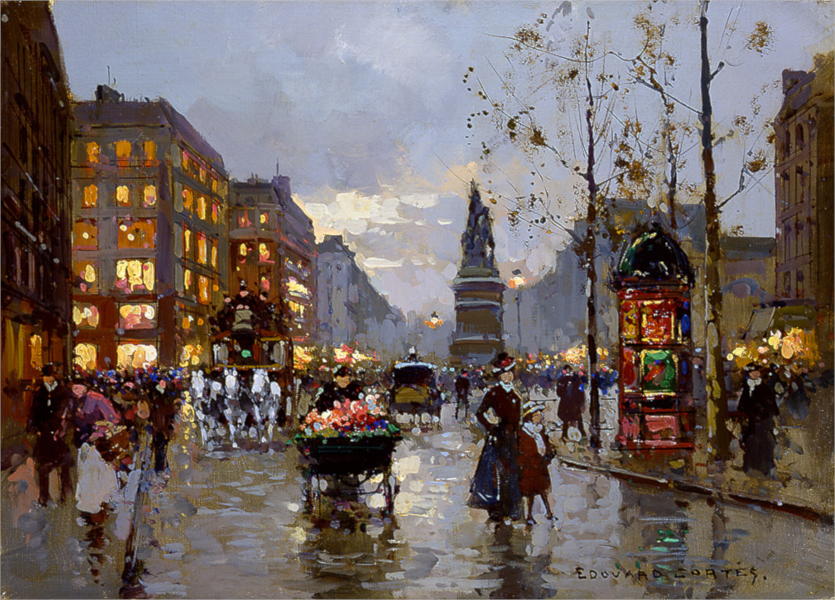} 
    
    \includegraphics[width=0.19\textwidth]{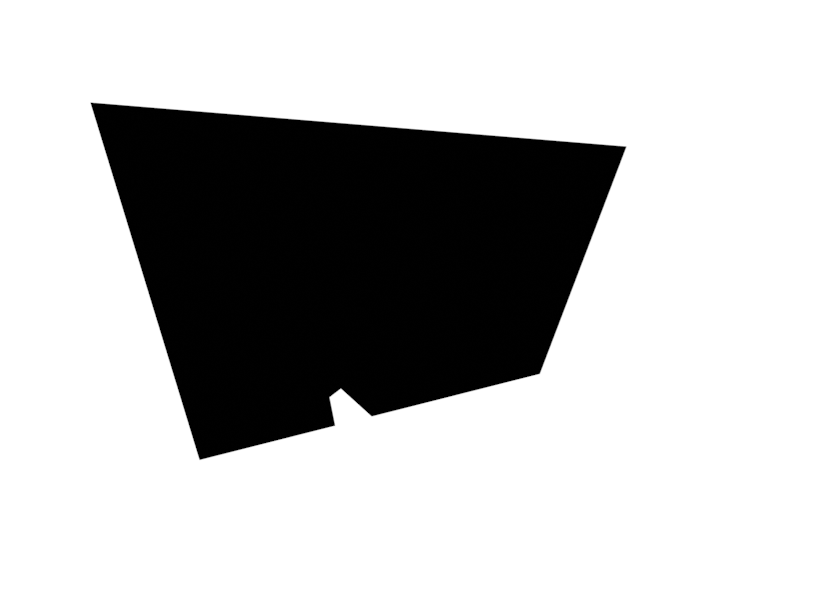}
    \includegraphics[width=0.19\textwidth]{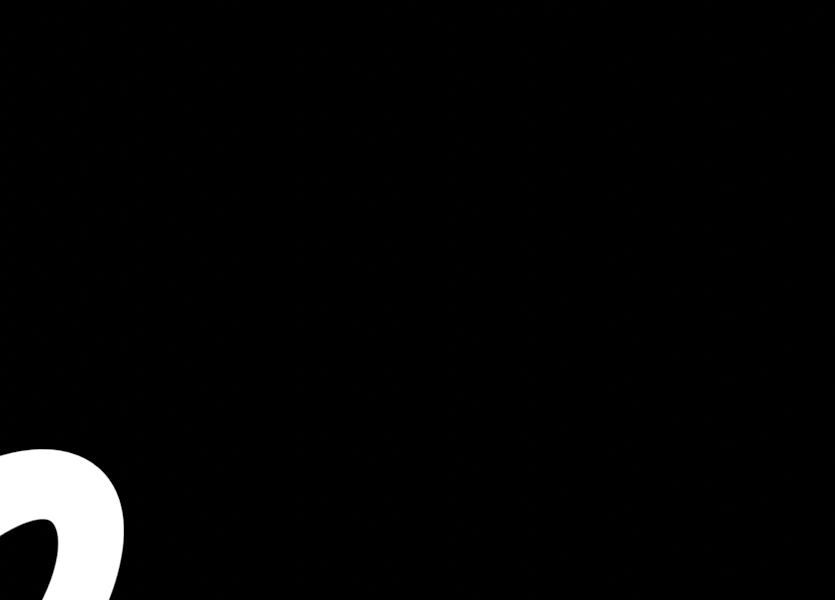}
    \includegraphics[width=0.19\textwidth]{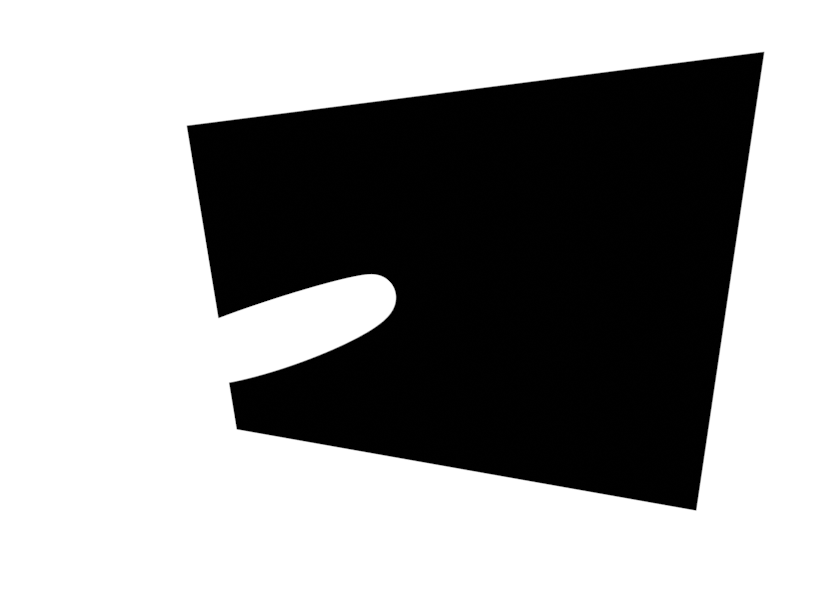}
    \includegraphics[width=0.19\textwidth]{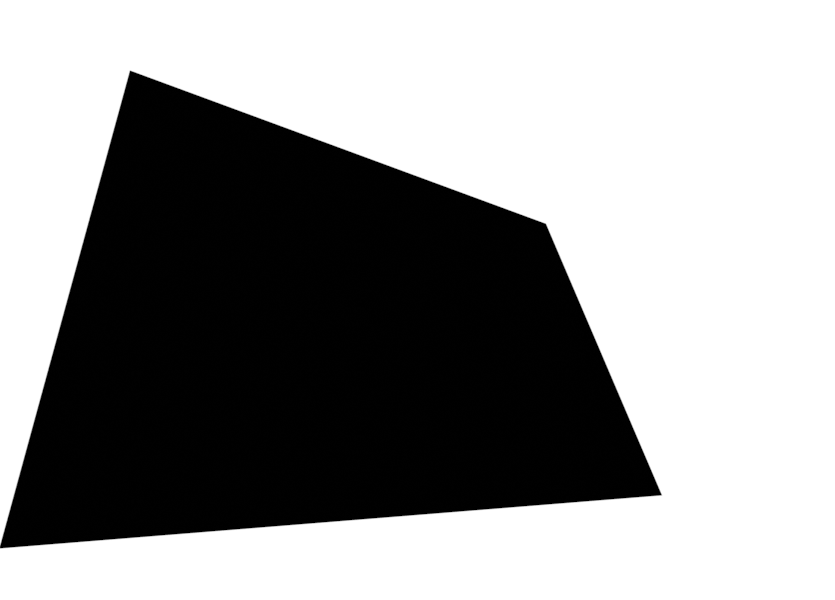}
    \includegraphics[width=0.19\textwidth]{figs/align/seq/0-m.png}
    \caption{The top row shows four generated images with perspective distortions and their corresponding label. The bottom row shows the occlusion mask for each image.}
    \label{fig:misaligned}
\end{figure}

\begin{figure}[h]
    \centering
    \includegraphics[width=\textwidth]{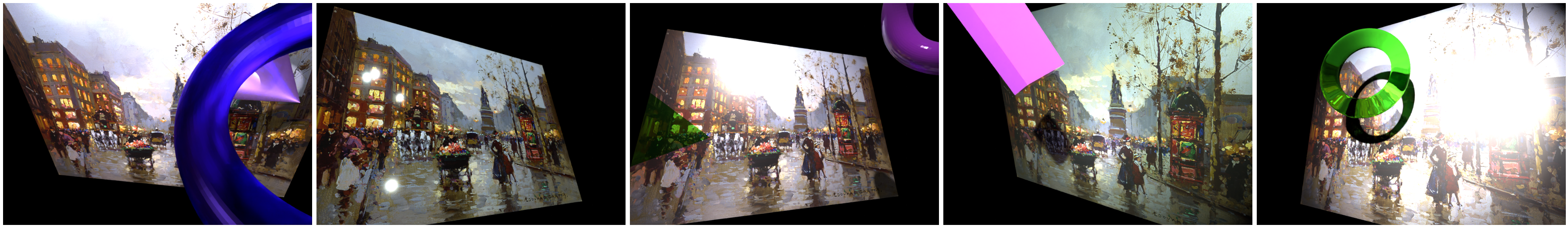}
    \includegraphics[width=\textwidth]{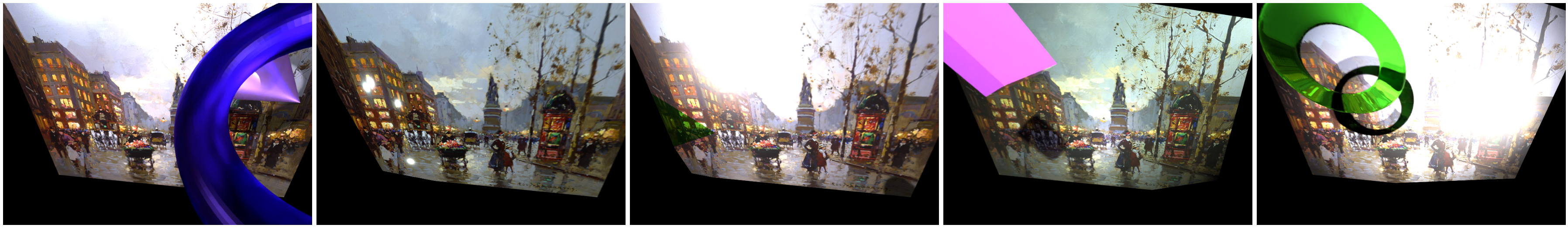}
    \caption{The first row shows a sequence of images with perspective distortions. The second row shows the images warped into the reference frame of the first image.}
    \label{fig:seq_align}
\end{figure}

In addition to the aligned dataset, we generate images with perspective distortions. For each randomly generated camera, we render an image. In order to evaluate the image alignment with reconstruction, we also generate a single ground truth image as described in section \ref{seq:aligned} under ambient lighting conditions.
Figure \ref{fig:misaligned} shows a sequence of distorted images. The last image contains no distortions and is aligned with the camera's field of view. \\
For each image pair $(I_i$, $I_j)$, the corresponding homography $H_{ij}$ is computed using DLT as described in section \ref{sec:dlt}. It is possible to warp any image $I_j: \mathbb{R}^2\mapsto \mathbb{R}^3$ into the reference frame of any other image $I_j: \mathbb{R}^2\mapsto \mathbb{R}^3$ using a warp function $\mathcal{W}_{ij}: \mathbb{R}^2\mapsto \mathbb{R}^2$. 
Figure \ref{fig:seq_align} shows a sequence of images under perspective distortion. Using the estimated homographies, all images can be aligned with the first image. 

 Homography Adaption is a technique that applies random homographies to images from MS COCO \cite{lin2014microsoft} for interest point detection and descriptor learning \cite{detone2018superpoint}.
 Although this technique is successfully applied in various image-matching tasks, it does not explicitly enforce invariance to illumination changes or occlusions. Our dataset allows to overcome these shortcomings. 
 Although our misaligned dataset also creates randomized homographies, it generates homographies from camera projections, which better represent realistic 2D projections.
 
The dataset containing perspective distortions extends all tasks mentioned in section \ref{sec:aligneddata}, but it also creates new challenges and applications. The following chapters discuss some potential applications of our dataset. 
\subsubsection*{Homography Estimation}
Our dataset provides the ground-truth homography for any image pair within a sequence, making it possible to train and evaluate deep homography estimation methods. Let $I_i$ and $I_j$ be two images, let $H_{ij}$ be their corresponding perspective transformation, and let $f_{\theta}(I_i,I_j)\in \mathbb{R}^9$ be a deep learning model that estimates a homography from two images. The learning objective can be described by a regression loss $L(\theta)$: 
\begin{align}
    \Hat{H_{ij}} &:= f_{\theta}(I_i,I_j) \notag \\
    L(\theta) &= \left\lVert \frac{1}{\sqrt[3]{|\hat{H_{ij}}|}}\hat{H_{ij}} - \frac{1}{\sqrt[3]{|H_{ij}|}}H_{ij} \right\rVert_F \label{eq:deep-homography}
\end{align}
Here $|\cdot|$ describes the determinant, and $\lVert \cdot \rVert_F$ describes the Frobenius norm. Both matrices have to be normalized since homographies are equivalent under a scaling factor. The given normalization fixes the scale of both matrices such that their determinants are equal to one. The objective functions allow the training of an end-to-end deep learning model. The data generation provides an indefinite amount of data. The changes in illumination and occlusion provide additional challenges and require a more robust estimation of features.

\subsubsection*{Bundle Adjustment}

Many existing homography estimation methods compute the alignment from image pairs only. This can be extended to sets of images. The problem can be described as a bundle adjustment problem. The SIDAR dataset can be used for neural bundle adjustment methods, such as BARF\cite{lin2021barf}. It could be possible to learn neural priors for bundle adjustment. Larger image sets also enforce more consistency across images compared to image pairs. 

\subsubsection*{Descriptor Learning}

Given the correspondences between images, local descriptors can be learned. Correspondences exist even under very strong distortions, which allows the development of descriptors that are invariant or equivariant to the given perturbations. The methodology of HPatches \cite{hpatches_2017_cvpr} could also be extended to the SIDAR dataset to add more variety in image patches.  

\subsubsection*{Dense Correspondences}

The ground-truth homographies also provide dense correspondences between each image point with high accuracy. Correspondences can be estimated not only for sparse key points but for every pixel. The neighborhood of points remains mostly unchanged under perspective distortions. This puts additional constraints on image-matching tasks. The occlusion masks also provide regions of outliers, while the other distortions add robustness. Image matching models can be trained to densely detect image regions under various perturbations and also detect outliers as points with no matches.

\newpage
\section{Conclusion}

In this work, we propose a data generation with a corresponding dataset based on 3D rendering that introduces various disturbances, such as shadows, illumination changes, specular highlights, occlusions, and perspective distortions, to any given input image. Although it is a synthetic dataset, the data augmentations are not trivial, and they are customizable. To the best of our knowledge, we provide the first large-scale dataset containing ground-truth homographies of image sequences for deep homography estimation, which does not consist of trivial perspective distortions. Our rendering pipeline allows us to both generate new datasets and augment existing data. Models trained on these augmentations can learn a representation that is robust to these perturbations. It can be used as an extension of Homographic Adaption \cite{detone2018superpoint} or contrastive learning \cite{chen2020simple}.
We discuss a range of computer vision applications for which this dataset can be used. It can contribute to the training of end-to-end deep learning models that solve image alignment and restoration tasks such as deep homography estimation, dense image matching, descriptor learning, 2D bundle adjustment, inpainting, shadow removal, denoising, content retrieval, and background subtraction.

The limitation of most synthetic datasets lies in their deviation from real data. This can result in biased models and limit generalization and, thus, model performance. Compared to existing augmentation methods that apply randomized homographies \cite{detone2018superpoint} to images, SIDAR adds additional complexity. Especially adding illumination changes, shadows, and occlusions can improve the robustness of learned descriptors and feature matching.

Future work could focus on developing benchmarks to provide specific evaluation metrics for the discussed tasks and compare the generalization across different datasets. 
Additionally, future work could further improve the realism of the data generation and add new data modalities, such as videos. 
The data generation can be adapted to include other distortions, such as reflective surfaces, translucent occlusions, or camera lens distortions. Many of these effects are common artifacts in real imaging systems, but it is difficult to create large-scale datasets for these cases. Our data generation can provide an effective way to approximate these artifacts and provide a large enough dataset for training and evaluation.
It can serve as a baseline to study these effects in a more controlled environment.

%\section{Benchmark}

\bibliographystyle{unsrt}  
\bibliography{references}  

\end{document}